\documentclass[conference]{IEEEtran}
\IEEEoverridecommandlockouts

\usepackage{algorithm}
\usepackage{romannum}
\usepackage{caption}
\usepackage{subcaption}
\usepackage{multirow}
\usepackage{booktabs}
\usepackage{cite}
\usepackage{amsmath,amssymb,amsfonts}
\usepackage{graphicx}
\usepackage{textcomp}
\usepackage{xcolor}
\usepackage{hyperref}
\usepackage{algpseudocode}
\usepackage[misc]{ifsym}

\def\BibTeX{{\rm B\kern-.05em{\sc i\kern-.025em b}\kern-.08em
    T\kern-.1667em\lower.7ex\hbox{E}\kern-.125emX}}
\begin{document}

\title{Tell Me How to Survey: Literature Review Made Simple with Automatic Reading Path Generation\\
}

\author{\IEEEauthorblockN{Jiayuan Ding$^{*\dagger}$}
\thanks{* Equal contributions.}
\thanks{$\dagger$ The work was done during internship at Mila.}
\IEEEauthorblockA{
\textit{Michigan State University}\\
dingjia5@msu.edu}
\and
\IEEEauthorblockN{Tong Xiang$^{*\dagger}$}
\IEEEauthorblockA{
\textit{Virginia Tech}\\
xtong@vt.edu}
\and
\IEEEauthorblockN{Zijing Ou$^*$}
\IEEEauthorblockA{
\textit{Sun Yat-sen University}\\
zijingou.mail@gmail.com}
\and
\IEEEauthorblockN{Wangyang Zuo}
\IEEEauthorblockA{
\textit{Zhejiang University of Technology}\\
zuowangyang@foxmail.com}
\and
\IEEEauthorblockN{Ruihui Zhao}
\IEEEauthorblockA{
\textit{Tencent Jarvis Lab}\\
zacharyzhao@tencent.com}
\and
\IEEEauthorblockN{Chenhua Lin}
\IEEEauthorblockA{
\textit{University of Sheffield}\\
c.lin@sheffield.ac.uk}
\and
\IEEEauthorblockN{Yefeng Zheng}
\IEEEauthorblockA{
\textit{Tencent Jarvis Lab}\\
yefengzheng@tencent.com}
\and
\IEEEauthorblockN{Bang Liu\textsuperscript{\Letter}}
\thanks{\Letter \ Corresponding author. Bang Liu is also affiliated
with Canada CIFAR AI Chair.}
\IEEEauthorblockA{
\textit{Université de Montréal $\&$ Mila}\\
bang.liu@umontreal.ca}
}


\maketitle

\begin{abstract}
Recent years have witnessed the dramatic growth of paper volumes with plenty of new research papers published every day, especially in the area of computer science.
How to glean papers worth reading from the massive literature to do a quick survey or keep up with the latest advancement about a specific research topic has become a challenging task.
Existing academic search engines return relevant papers by individually calculating the relevance between each paper and query. However, such systems usually omit the prerequisite chains of a research topic and cannot form a meaningful reading path.
In this paper, we introduce a new task named Reading Path Generation (RPG) which aims at automatically producing a path of papers to read for a given query. To serve as a research benchmark, we further propose SurveyBank, a dataset consisting of large quantities of survey papers in the field of computer science as well as their citation relationships.
Furthermore, we propose a graph-optimization-based approach for reading path generation which takes the relationship between papers into account. Extensive evaluations demonstrate that our approach outperforms other baselines. A real-time \textbf{Re}ading \textbf{Pa}th \textbf{Ge}ne\textbf{r}ation (\textbf{RePaGer}) system has been also implemented with our designed model.
Our source code and SurveyBank dataset can be found \texttt{here} \footnote{\url{https://github.com/JiayuanDing100/Reading-Path-Generation}}.

\end{abstract}


\renewcommand{\algorithmicrequire}{\textbf{Input:}}
\renewcommand{\algorithmicensure}{\textbf{Output:}}

\section{Introduction}\label{sec:intro}
With the explosion of the quantity of scientific papers, it becomes more challenging for researchers to glean papers worth reading from a large amount of literature for a quick review or keeping up with the latest advancement.
Given a research topic that a user is not familiar with, he/she may try to read some survey papers about the topic and refer to the reference list in the surveys to get more details.
However, a high-quality and up-to-date survey is not always available for any given research topic. A user may also utilize academic search engines
to find out related papers to read. Existing academic search engines usually retrieve a list of articles by calculating semantic similarity between user's query and each individual article separately, ignoring the prerequisite relationship among searching results.
Therefore, it will be highly valuable if we can retrieve a list of core papers with reading order for a given query to show users a reading path. In this way, a user can read the generated reading path and quickly get familiar with a new research field.

In recent years, research about scientific paper analysis has attracted a lot of attention. However, most existing works focus on citation recommendation~\cite{zhou2008learning, tang2009discriminative, he2010context, ren2014cluscite, sun2017recommendation}, which aims at recommending citations for a given text. Another category of research works focus on citation classification \cite{jurgens2016citation, teufel2006automatic, dong2011ensemble, zhao2019context, cohan2019structural, pride2020authoritative}, aiming to reason author's intention for citing a paper.
There are also research works about generating a reading list to users. \cite{jardine2014automatically} proposed ThemedPageRank to create reading lists by re-ranking topic-related papers. \cite{ekstrand2010automatically} built up a reading list for a given query set by exploring collaborative and content-based filtering techniques.
But these methods ignore the prerequisite relationship among the returned papers. \cite{gordon2017structured} designed a concept graph-based approach to generate structured reading lists, which takes into account the prerequisite concepts of the topic. However, this method strongly relies on the quality of existing concept graph.

Previous research about reading list generation mainly solves the problem of ``what to read" by producing a list of papers for a given research topic. However, to serve users unfamiliar with a research field, two critical problems are omitted by existing research: ``how to read" and ``how to understand".
For the first problem, we need to organize different research works in a structured manner, so that a user can read the papers according to the paths in the structure, and get a sense of how the research about a specific topic evolves during history. Note that simply organizing papers by publication time is not enough, as the correlation between different research is overlooked.
For the second problem, we need to include some papers in our generated path which may be not directly related to the queried research topic, but are quite helpful or essential to understand the highly related papers in our generated list. 

Motivated by above analysis, in this paper, we introduce a new task {\em Reading Path Generation (RPG)}, which aims to automatically create a reading path for a given query. 
Our task is designed to help researchers who are new to a field to do a quick survey by solving the two problems mentioned above.
Specifically, we are aiming to retrieve a list of papers which includes the research works highly related to the given query, as well as prerequisite research works that are essential to understand the highly related works.
Furthermore, we organize these papers to form a reading path, which takes both the chronological order and the correlation between different research works into account.
We hope that by reading the retrieved papers according to the reading path, a user can easily understand each research work, quickly complete a survey and form a landscape of the research history of any topic.

\begin{figure*}[!t]
\centering
\includegraphics[width=\linewidth]{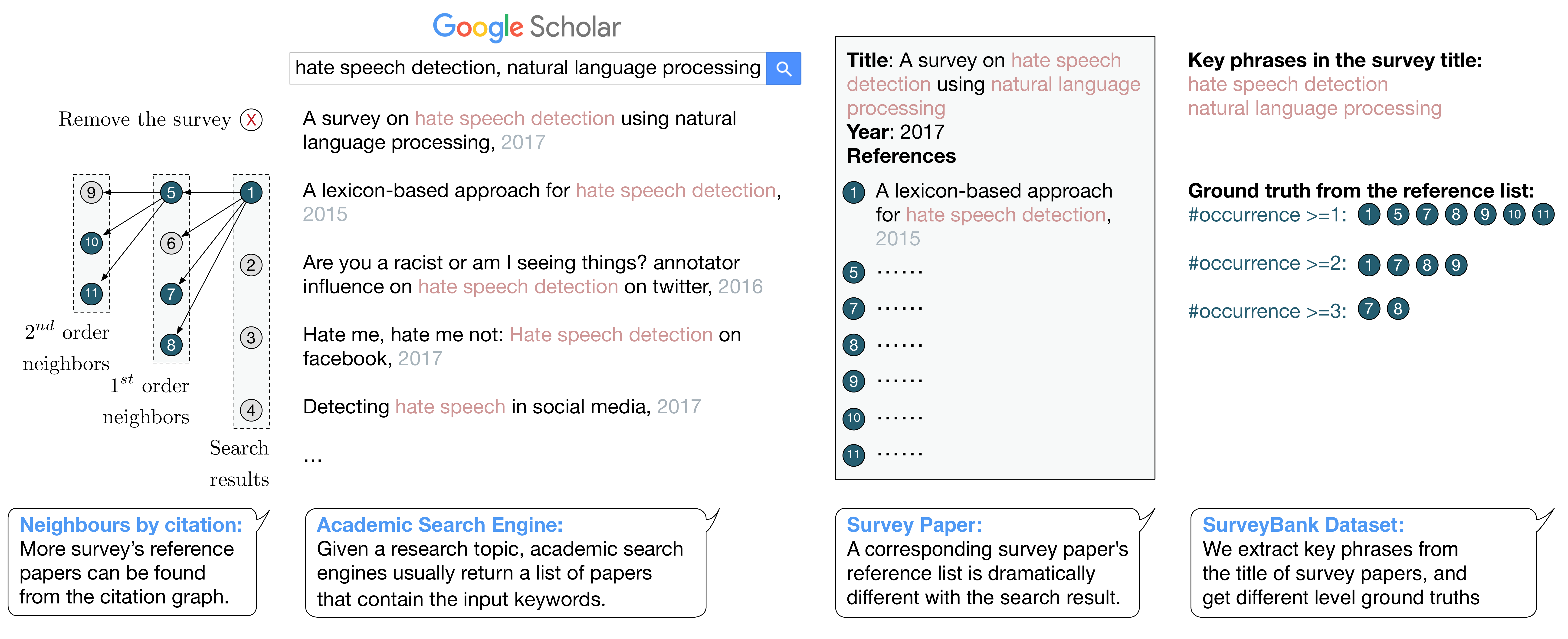}
\vspace{-4mm}
\caption{A real-world example of the comparison between Google Scholar and a survey paper. In this figure, each circle represents a paper, and dark ones are overlapping papers between the searching result of Google Scholar and a survey's reference list. An arrow between two papers indicates the reference relationship. For example, Paper 1 $\rightarrow$ Paper 5 means Paper 1 cites Paper 5. The right side in this figure illustrates the evaluation benchmark for our Reading Path Generation task. The number of occurrences at the right side indicates how many times a reference paper was cited in a survey.}
\label{fig:real-world-example}
\vspace{-4mm}
\end{figure*}

\begin{figure}[!t]
\centering
\begin{subfigure}[t]{0.236\textwidth}
\centering
  \includegraphics[width=\textwidth]{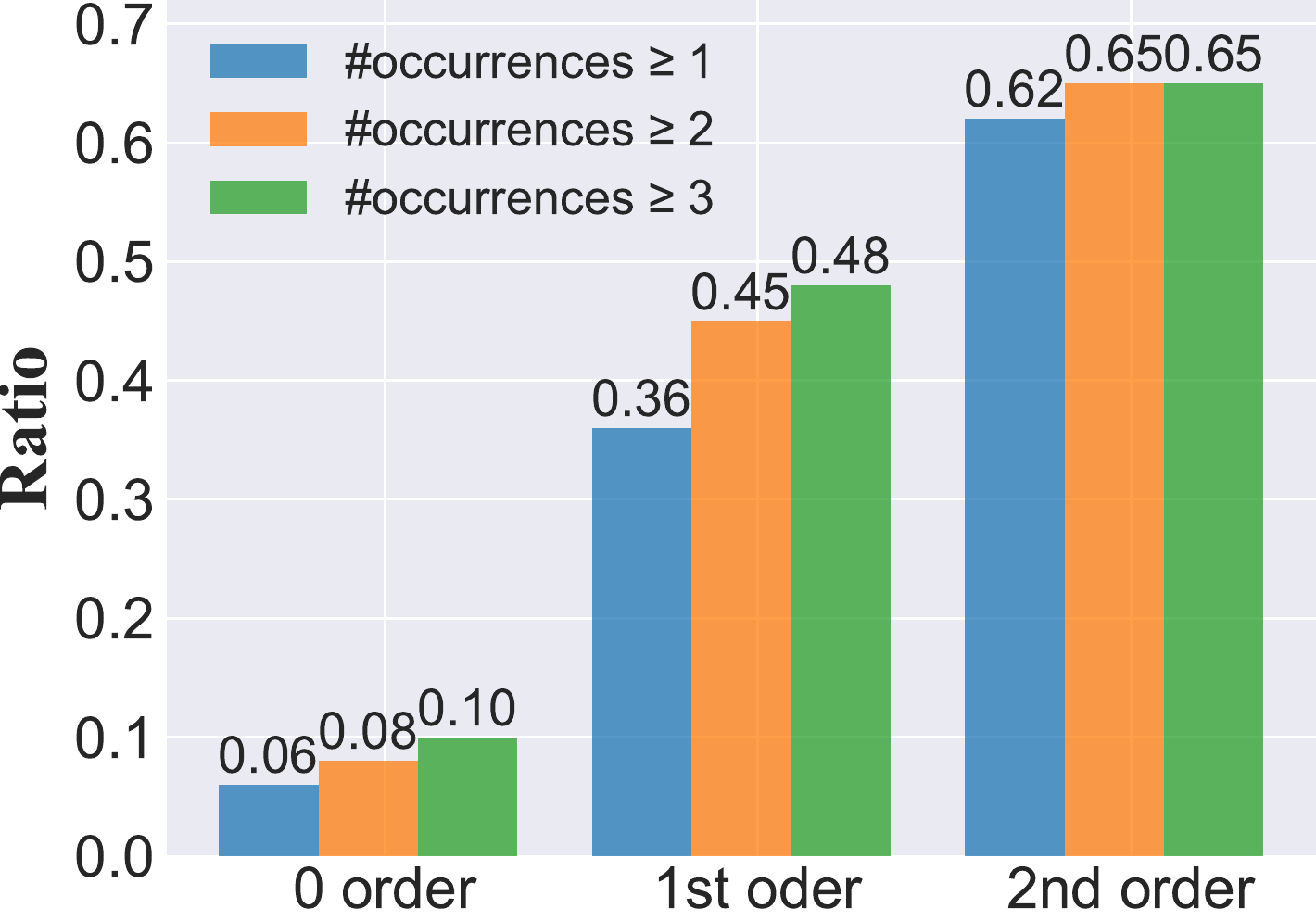}
  \caption{TOP 30}
\end{subfigure}
\begin{subfigure}[t]{0.236\textwidth}
\centering
  \includegraphics[width=\textwidth]{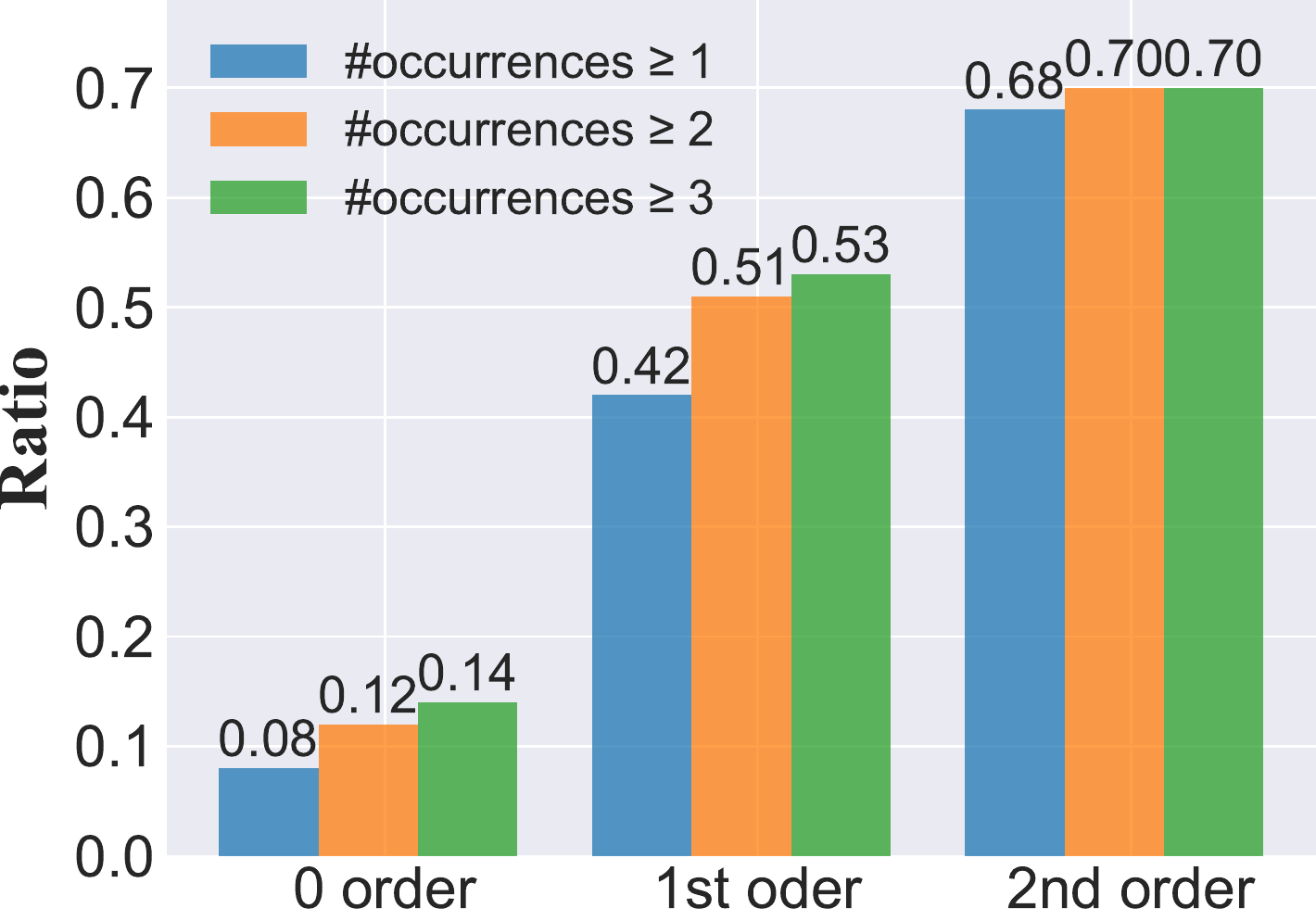}
  \caption{TOP 50}
\end{subfigure}
\caption{TOP30/TOP50 refers to the number of initial seed papers returned from Google Scholar}
\vspace{-4mm}
\label{fig:stats-comparison}
\end{figure}

Due to the lack of benchmark datasets for evaluating Reading Path Generation, we further propose the {\em SurveyBank} dataset which is specifically designed for our newly proposed task.
We create SurveyBank without any human labeling based on two key insights.
The first one is that a lot of existing high-quality survey papers are already available, and each of them is written by professional researchers, encompassing highly valuable information about how to form a survey about a research topic.
Based on this observation, we utilize the key phrases extracted from survey titles as the inputs of RPG task, and reading paths inferred from surveys' reference lists as true labels.
We initially collect $41,194$ surveys in the field of computer science from Google Scholar and S2ORC \cite{lo-etal-2020-s2orc}. After that, $9,321$ high quality surveys are selected to form SurveyBank. It covers 10 topics of computer science like Artificial Intelligence, Computer Network and Database. 
The second key insight is that the citation relationship between different papers reflects the coherence between different papers.
Therefore, we construct a large-scale scientific citation graph with 6 million papers from S2ORC~\cite{lo-etal-2020-s2orc} in the field of computer science by citation relationship, and this citation graph covers reference lists of each survey.
To the best of our knowledge, SurveyBank is currently the largest survey dataset. 

Last but not least, we also propose and implement a Real-time Reading Path Generation System (RePaGer)
which takes key phrases as inputs and generates a reading path of papers for users to read. 
In RePaGer, we start by getting back top-$K$ articles called initial seed papers from Google Scholar. After that, we re-allocate seeds by first capturing neighbors of initial seed papers as seed candidates and then selecting articles with high co-occurrence.
We infer that these new selected seeds are more likely to be prerequisite concepts of a topic since multiple articles directly relevant to the topic cite them in their papers. 
Finally we propose Node-Edge Weighted Steiner Tree (NWST), a graph optimization-based approach, to identify a reading path with prerequisite chains for a given query. 
Our model not only considers the importance of each paper in a whole citation network but also takes into account the relationships among them. 

We run extensive evaluations based on the SurveyBank dataset and compare our proposed approach for Reading Path Generation with existing academic search engines and baseline methods. The evaluation results demonstrate that our approach outperforms other baselines and can generate insightful reading path for different research topics. 


\section{TASK DEFINITION}\label{sec:task-defi}
In this section, we compare the articles from Google Scholar and a corresponding survey's reference list to motivate our work. We then formalize our proposed new task.

\subsection{Background and Motivation}
Given a research topic, people usually rely on academic search engines such as Google Scholar to retrieve a list of papers to read. Our key observation is that the paper list returned by search engines has dramatic difference with the reference papers in a survey for the same topic. This is explained by the fact that the returned ranked list from such search engines only considers relevance with the queried topic but not includes prerequisite chains of the queried topic. However, a reference list in a survey paper is comprehensive to cover more concepts including prerequisite chains or sub-topics of the queried subject.

\textbf{Experimental Setup} Figure \ref{fig:real-world-example} illustrates an example of comparing the searching results from Google Scholar and the reference list from a survey paper \cite{schmidt2017survey}.
The query in Google Scholar is the key phrases extracted from a survey title. 
For searching setup, the time range is set from anytime to the year when this survey is published so that articles published after the survey will not be returned. We also remove the survey paper itself if found in the result of Google Scholar. 
To avoid contingency, we select a subset of SurveyBank (details in Section \ref{sec:dataset}) with high scores $s$ for further statistics, where the score is calculated as follow: 
\begin{equation*}
s = \frac{citation}{2020-year+1} 
\end{equation*} 
where \textit{citation} refers to the number of citations of a survey paper and \textit{year} is when the survey is published.



\textbf{Evaluation} Because the importance of each paper in a survey's reference list is not the same, as on the right side in Fig. ~\ref{fig:real-world-example}, we create three labels for each survey based on how many times a referred paper is mentioned (or cited) in the survey: a full reference list (\#occurrences $\geq$ 1), a reference list where each paper is cited at least twice (\#occurrences $\geq$ 2) and a reference list where each paper is cited at least three times (\#occurrences $\geq$ 3).
For top 30 and top 50 searching results from Google Scholar, we respectively count the ratio of overlapping papers between scholar response of $0^{th}$ order, 1$^{st}$ order, 2$^{nd}$ order and the three reference lists.

\textbf{Observations} From Fig. \ref{fig:real-world-example}, we can observe that only one paper shown in the top $5$ result belongs to the survey's reference paper list (i.e., Paper 1). From Fig. \ref{fig:stats-comparison}, we can also notice that the overlapping ratio between initial retrieved articles from Google Scholar and the survey's reference paper list is pretty low no matter it is TOP 30 or TOP 50. 
However, if we further explore the papers cited by the top $K$ result of Google Scholar ($1^{st}$ order neighbors), more survey's references will show up (i.e., Paper 5, 7, and 8). We can repeat this operation by further exploring the papers cited by the $1^{st}$ order neighbor papers (i.e., the $2^{nd}$ order neighbors), then more papers belonging to the survey's references would be found (Paper 10 and 11).
It is also proved from Fig. \ref{fig:stats-comparison} that if we further explore 1$^{st}$ order and 2$^{nd}$ order neighbors of initial TOP 30/50 seed papers, the overlapping ratio will greatly increase. To conclude, we can summarize these into two main observations below: 

\begin{itemize}
    \item \textbf{Observation I:} For the same topic, the articles directly returned from Google Scholar have a huge gap with the reference list of a survey. 

    \item \textbf{Observation II:}  Although there is a gap between the two, most of the missing papers can be retrieved through the neighbors (1$^{st}$ order and 2$^{nd}$ order citations) of the initial seed nodes (i.e., Google Scholar results). 
\end{itemize}

\textbf{Understandings} In this subsection, we provide the understanding and explanation for aforementioned observations.
\begin{itemize}
    \item \textbf{Understanding of Observation I:} The results from Google Scholar are only directly related to the topic of the query, while a survey is comprehensive to cover more concepts including prerequisite chains or sub-topics of the subject.
    
    \item \textbf{Understanding of Observation II:} In each scientific paper, pre-requisite work or sub-topics of the object can be mentioned beforehand in related work section for introductory purpose. It is reasonable that pre-requisite work or sub-topic papers can be found in the neighbors of initial seed papers.
    
\end{itemize}

The observations and understandings above motivate us to design a new task called Reading Path Generation, which not only returns papers highly relevant with the queried topic but also retrieve prerequisite papers to form a path or paths to help better understand the topic or those highly relevant papers. The observations also motivate the design of our graph-optimization-based approach to solve this problem in Section \ref{sec:model}.



\subsection{The Task of Reading Path Generation}
Motivated by above observations, we propose the task of Reading Path Generation. Here we first introduce some necessary definitions, and then formally formulate our task.

\textbf{Evaluation Benchmark}. The evaluation benchmark for our task consists of a set of surveys $\mathcal{D} = \{ d_1, d_2, ..., d_{|\mathcal{D}|} \}$. 
As shown in Fig.~\ref{fig:real-world-example}, each survey $d$ is accompanied by a set of key phrases $\mathcal{K}=\{ k_1, k_2, ..., k_{|\mathcal{K}|} \}$ extracted from its title, as well as a set of true labels $\mathcal{V}$ inferred from its reference list. 
For each survey, since the importance of each paper in a survey’s reference list is not the same, we design multiple true labels in $\mathcal{V}$ by counting how many times a reference paper was cited in a survey: $\mathcal{V} = \{ \mathcal{L}_1, \mathcal{L}_2, \mathcal{L}_3 \}$, where $\mathcal{L}_i$ represents a list of papers where each was cited at least $i$ times in the survey.

\textbf{Reading Path Generation Task}. Given input key phrases $\mathcal{K}$, our goal is to learn a mapping function $\mathcal{F}$: $\mathcal{K} \longrightarrow \mathcal{O} $ to generate a reading path which covers the reference papers $\mathcal{V}$ as much as possible, where $\mathcal{O}$ is a list of generated papers with reading order. The reading order in our task refers to the direct/indirect citation relationships between papers.

\subsection{Task Evaluation}
How to define a reasonable reading path for a list of papers is an open question, and it is not accurate to evaluate whether the specific reading path is correct or not with metric evaluation.
Because for a given list of papers, there may be many optionally reasonable reading paths to be organized.
In this paper, we present that the citation relationship is one of the intuitive and reasonable ways to generate the structured reading path for the given list of papers. Of course, it is encouraged to incorporate other useful information to generate a more ingenious reading path, we leave it for future work. 
Therefore, we divide the evaluation on the generated reading path into two steps: 

i) \textbf{Overlapping metric evaluation on the reading list;} 

ii) \textbf{Human evaluation on the generated reading path.} 

\textbf{Why not metric evaluation like MAP on the ranked list instead of overlapping metric evaluation on the reading list?}
It is not possible to define a reading path as a ranked list $\{ P_1, P_2, ..., P_n\}$ where the paper at the front of the list has higher importance to the queried topic. Because a sequence of papers in the reading path only indicate the reading order instead of the importance of each paper, and it is hard to differentiate the importance of each paper in the reading path as well. Considering these, we evaluate the flattened reading path with overlapping metrics.

\textbf{Can the citation relationship between papers determine unique reading order?}
Yes, it can.
It’s hard to define the ground-truth reading order given a reading list. Different experts may have different opinions. 
We propose that the citation relationship combined with publish time is one of intuitive and reasonable ways to generate the structured reading path for the given list of papers. Once the reading list is determined, the reading direction between two papers can be easily and uniquely obtained from our constructed citation graph based on citation relationship and published time. For the case of multiple citation paths between two papers, we will assign all paths to them.
It can be explained by the fact that there can be multiple sub-topics related with the queried topic.

\section{The SurveyBank Dataset} \label{sec:dataset}
Here we introduce the \textit{SurveyBank} dataset as the benchmark for the evaluation of RPG task. We first present the overview of the dataset and the construction process, and then highlight the statistical properties of SurveyBank. Moreover, we briefly introduce the building procedure of the citation graph.

\subsection{Overview of SurveyBank}
Our goal of building SurveyBank is to provide a clean and scalable benchmark for RPG. Some previous works such as S2ORC~\cite{lo-etal-2020-s2orc} have been focusing on building a large-scale linked scientific corpus. These datasets cannot be directly transformed to RPG task since they \romannum{1}) mainly focus on citation graph, and \romannum{2}) suffer from low coverage of full text, disarray of paper types as well as topic domains. Other works~\cite{gordon2017structured} were focusing on tasks which are similar to RPG, but they require the assistance of domain experts which potentially lead to low scalability and additional human efforts. To mitigate the shortcomings of existing systems mentioned above and to better serve the task of RPG, we propose SurveyBank, a dataset constructed from a collection of survey papers without human labeling, as our benchmark. Each paper in it has its corresponding parsed full context containing hierarchical section information, key phrases extracted from the title and ground truth labels from the references.

Our dataset also contains the input key phrases and ground truth labels to better serve as a benchmark for RPG. As we have mentioned in Section~\ref{sec:task-defi}, RPG takes one or more keywords/key phrases as its input. Since only a few surveys contain keywords as an inherent part of their papers, we choose to infer keywords/key phrases from the titles since titles can indicate the topics of the paper in many cases~\cite{jardine2014automatically}. 
For ground truth labels (the list of papers for generating desired reading path), we start with calculating the number of occurrences for each paper cited in the survey. The intuition behind this is simple: the more a paper is cited within the survey, the more likely this paper should be considered as significant. 
As described in Section~\ref{sec:task-defi}, we use generated lists to denote different significance levels for cited papers in the survey.

\subsection{Dataset Construction}

\begin{figure}[t]
\centering
  \includegraphics[width=\linewidth]{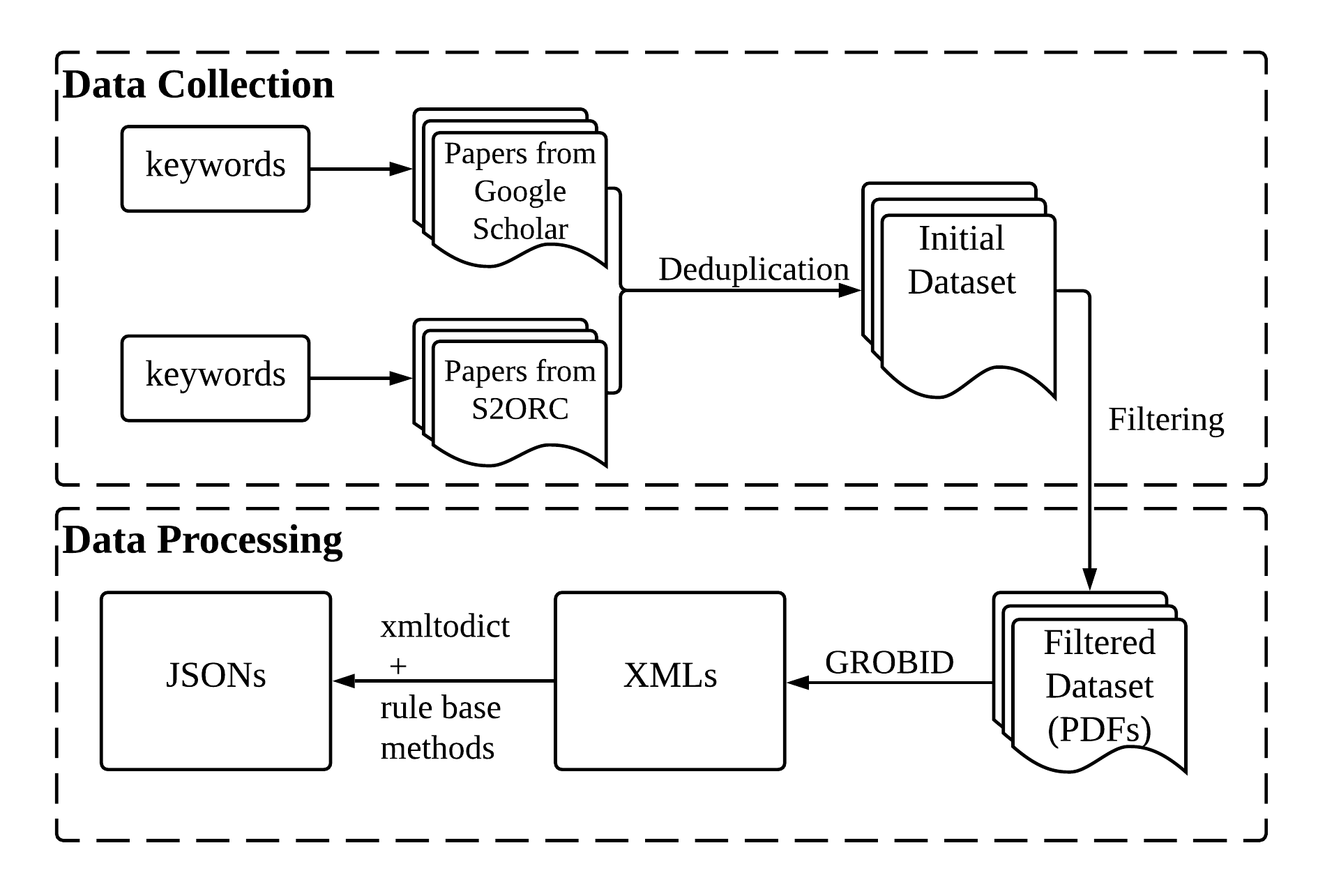}
  \caption{Overview of the dataset construction process.}
  \label{fig:flow diagram}
  \vspace{-4mm}
\end{figure}

The overall process for constructing SurveyBank is shown in Fig.~\ref{fig:flow diagram}. Initially, we collect survey papers based on keywords that indicate whether the paper is a survey or not, as well as topics to restrict the domain of the surveys collected. Our papers come from two sources: Google Scholar and S2ORC~\cite{lo-etal-2020-s2orc}. To make sure that the collected surveys are in the domain of computer science, we aggregate topic concept phrases from LectureBank~\cite{li2019should} and TutorialBank~\cite{fabbri-etal-2018-tutorialbank}. LectureBank provides $208$ topic keywords across $5$ different sub-domains in computer science, including Natural Language Processing (NLP), Machine Learning (ML), Artificial Intelligence (AI), Deep Learning (DL), and Information Retrieval (IR), verified by experts. On the other hand, extra $306$ topic keywords are collected from TutorialBank. These topic keywords are collected according to several criteria and also performed by experts, where the goal is to make the topic keywords more conceivable. We mix these topic keywords together and apply deduplication process to them, leaving us $441$ unique topic keywords. We utilize them together with additional keywords such as ``survey'' as inputs to collect survey papers from Google Scholar. For S2ORC, we retrieve the papers containing survey-indicating keywords such as ``survey'' in their titles from the computer science subset. Finally, we obtain an initial collection with $41,194$ papers in total. We further check paper titles in order to make sure there is no duplication among them and discard samples without proper full text.

We process the surveys' full text in PDF format following a similar procedure as S2ORC~\cite{lo-etal-2020-s2orc}. Specifically, \texttt{GROBID}~\cite{GROBID} is utilized to extract metadata, body text, and bibliography entries from PDFs into the XML format. In addition, we extract the keywords/key phrases from the title using the TopicRank algorithm~\cite{bougouin-etal-2013-topicrank} implemented by \texttt{pke}~\cite{boudin:2016:COLINGDEMO} and provide fine-grained structure for body text for each survey. The atomic units here are paragraphs instead of sections; for each section or subsection with a proper headline, we assign a notation to it to denote its hierarchical position within the survey. We hope that other researchers can make further progress in RPG task by leveraging the paper structure information. A survey is excluded from SurveyBank if it meets one of the following criteria: \romannum{1}) the corresponding PDF full text cannot be processed by the Python library \texttt{PyPDF2}~\cite{PyPDF2}; 
and \romannum{2}) the survey is more than 100 pages or less than 2 pages (A paper with more than 100 pages is more likely to be a thesis or reports). These parsed XMLs are then transformed into the JSON format; we apply a rule-based methods to the data processing procedure in addition to the Python library \texttt{xmltodict}~\cite{xmltodict} in order to mitigate certain inherent errors caused by \texttt{GROBID} and \texttt{xmltodict}. Our SurveyBank dataset is publicly available and can be found \underline{\href{https://github.com/JiayuanDing100/Reading-Path-Generation}{{\textcolor{blue}{here}}}}.

\subsection{Statistical Property}

\begin{figure*}[t]
\centering
\begin{subfigure}[t]{0.32\textwidth}
\centering
  \includegraphics[width=\textwidth, height=4.2cm]{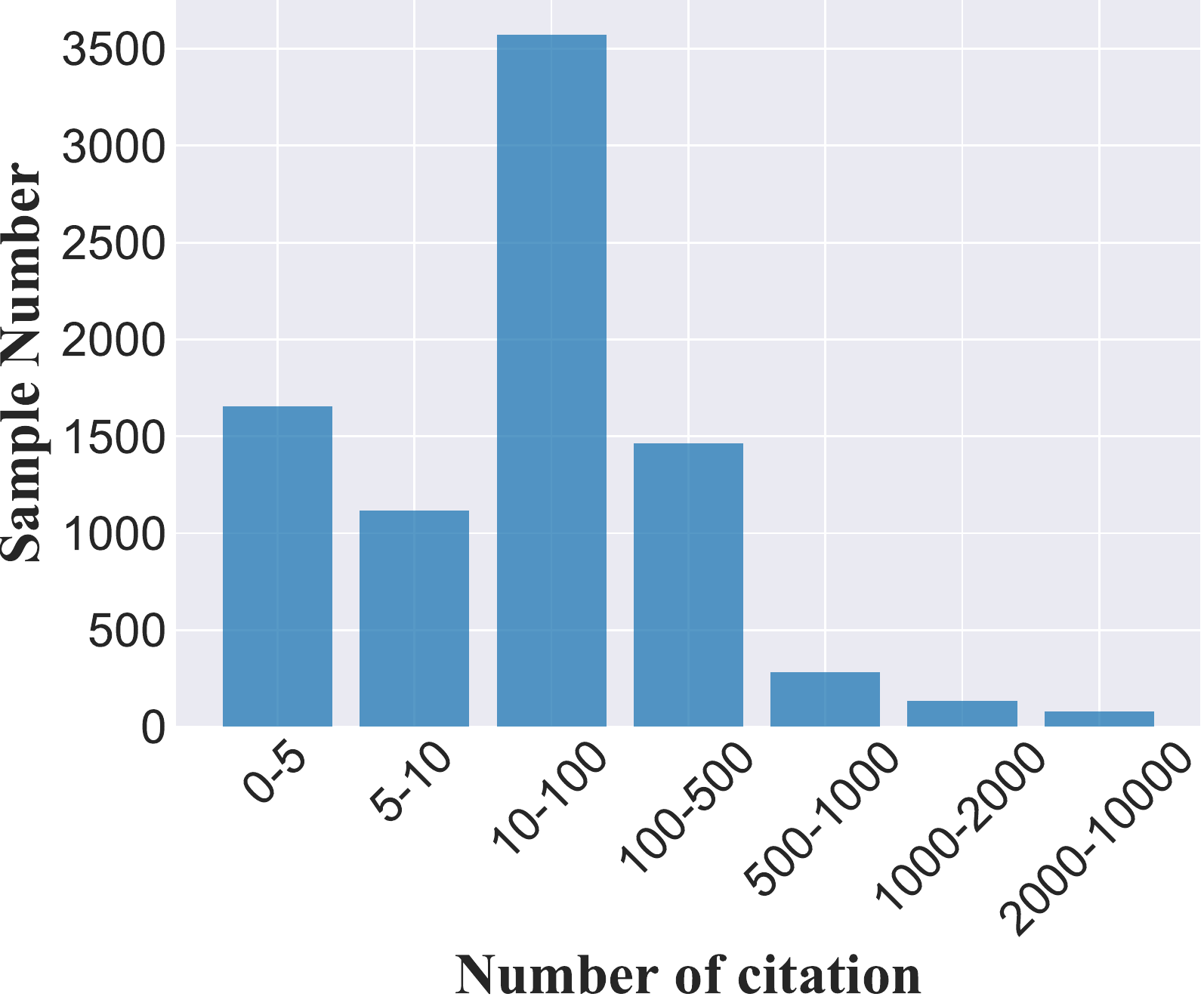}
  \caption{Distribution of the citation numbers of the survey papers in SurveyBank.}
  \label{cite_dis}
\end{subfigure}
\hfill
\begin{subfigure}[t]{0.32\textwidth}
\centering
  \includegraphics[width=\textwidth, height=4.2cm]{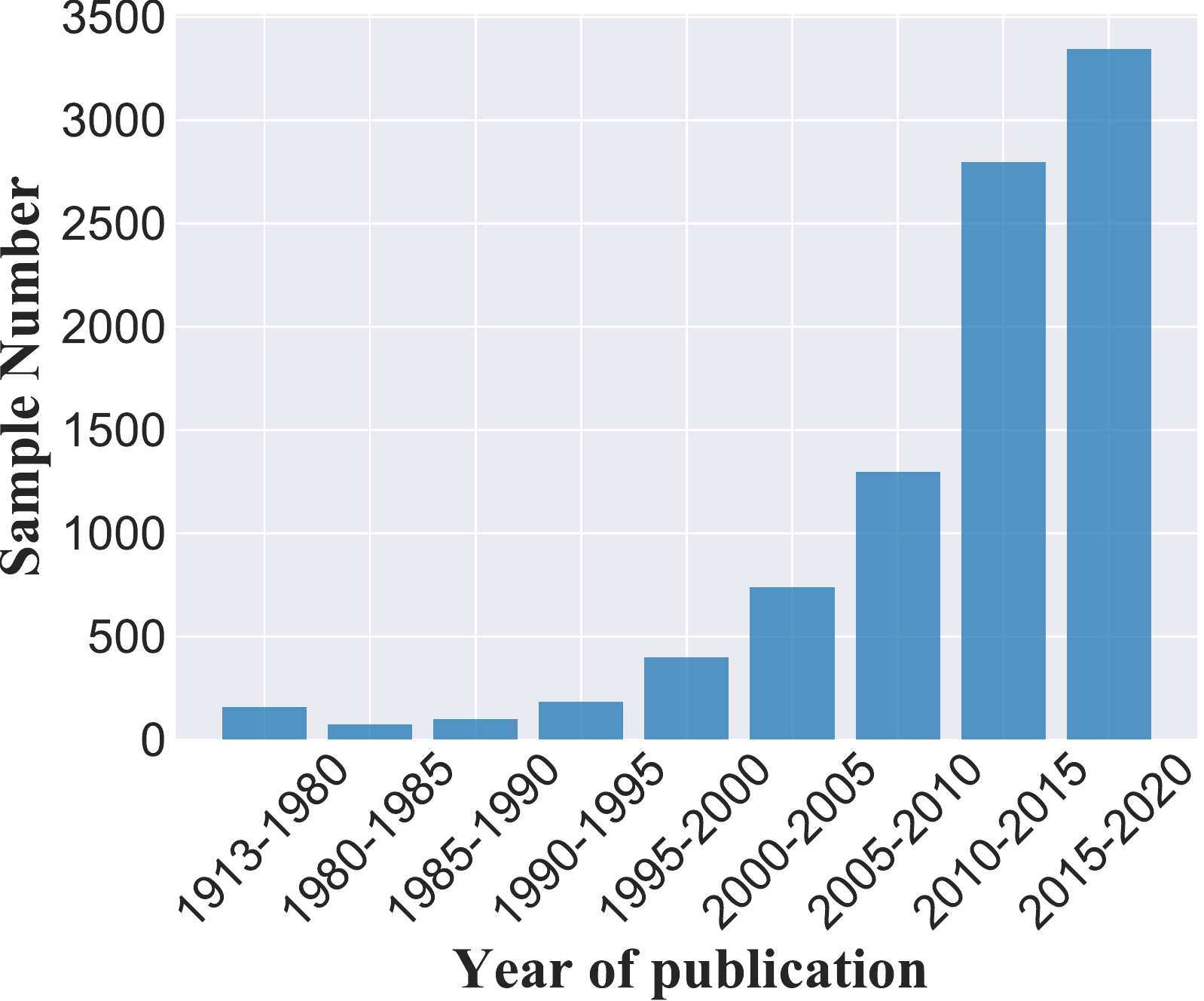}
  \caption{Distribution of publication years of SurveyBank.}
  \label{date_dis}
  
\end{subfigure}
\hfill
\begin{subfigure}[t]{0.32\textwidth}
\centering
  \includegraphics[width=\textwidth, height=4.2cm]{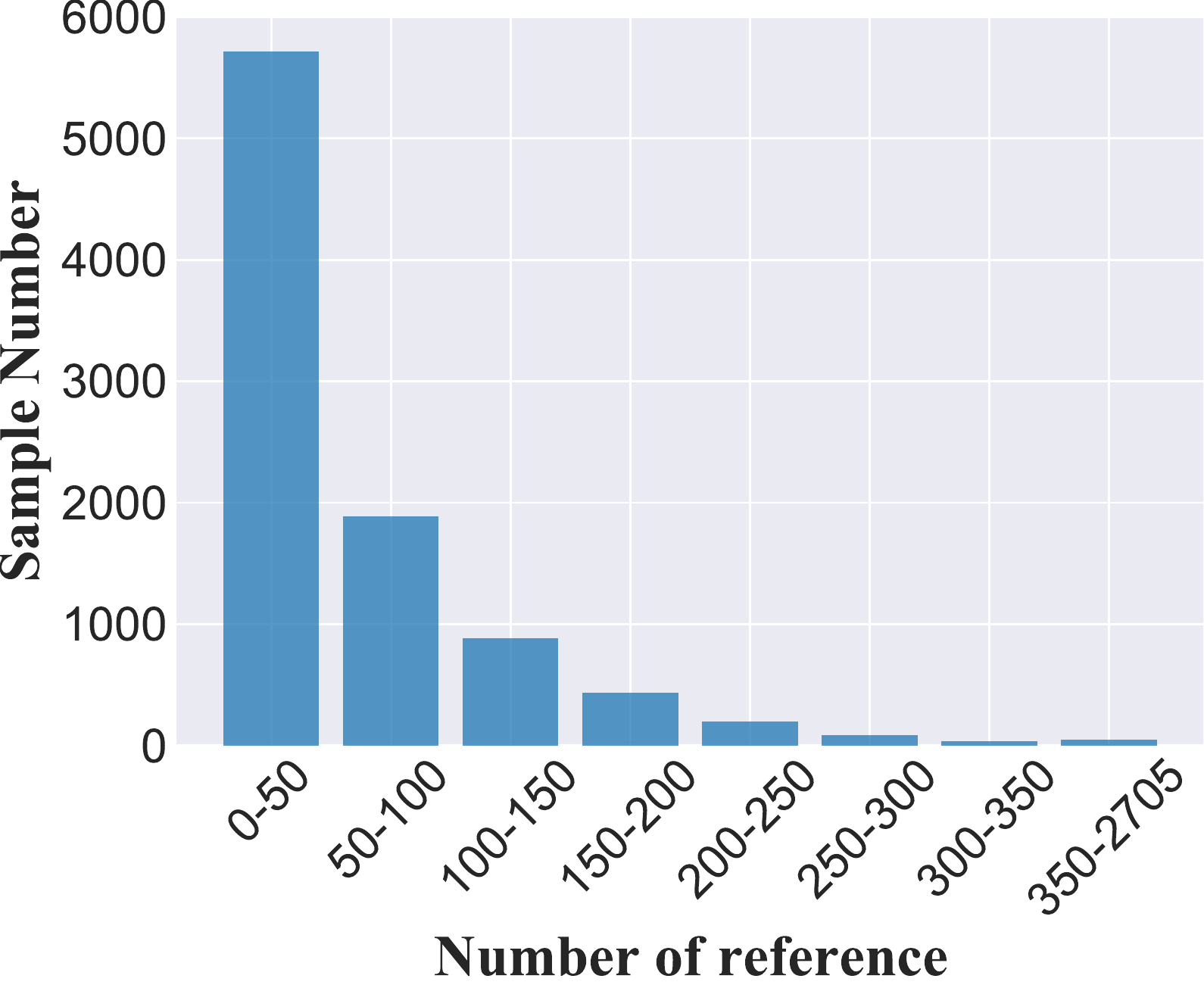}
  \caption{Distribution of the number of papers cited by survey papers in SurveyBank.}
  \label{bib_dis}
\end{subfigure}
\caption{Statistics of SurveyBank.}
\vspace{-5mm}
\label{fig:surveybank}
\end{figure*}

We calculate the statistics of SurveyBank, as shown in Fig. ~\ref{fig:surveybank}. Overall, SurveyBank contains $9,321$ survey papers with high quality filtered from an initial collection with $41,194$ survey papers. 
Each paper has approximately $58$ references on average.
One criterion for evaluating the impact of academic papers is how many times it has been cited overtimes. Intuitively, the more times one survey paper is cited, the more significant role the paper may be playing in its research domain throughout years of development.
To reflect this criterion, we also calculate the citation numbers for each survey paper within SurveyBank. Overall, $17.8\%$ of surveys have no opportunity of being cited by other papers, while $5.3\%$ of the surveys were cited for more than $500$ times over the years.
A detailed reference quantity distribution can be found in Fig. ~\ref{cite_dis}. In the case of publish date, more than $87.8\%$ of survey papers in SurveyBank are published within the recent $20$ years. The distribution of publication years and citation numbers for SurveyBank are shown in Fig. ~\ref{date_dis}. Eventually, we show the distribution of the number of papers cited by the survey papers in SurveyBank in Fig. \ref{bib_dis}.

Finally, to explore what kind of papers we have collected, we look into the domain distribution of papers in SurveyBank by checking where they were published according to a venue collection released by CCF~\cite{ccf}. The distribution is listed in Table~\ref{all domain distribution}. Here the category \textit{Uncertain Topics} contains the papers that either do not contain a valid publication venue inside its full text or their publication venues are not included in the venue collection provided by CCF.

\begin{table}[!t]
\footnotesize
\centering
\caption{Topic distribution of the survey papers in SurveyBank.}
\begin{tabular}{p{6cm}|c}
\toprule
\textbf{Domain} & \textbf{\#Papers}\\
\midrule
Artificial Intelligence & $1,151$ ($12.3\%$)\\\hline
Interdisciplinary, Emerging Subjects & $440$ ($4.7\%$) \\\hline
Computer Network & $424$ ($4.5\%$) \\\hline
Computer Graphics and Multimedia & $280$ ($3.0\%$) \\\hline
Database, Data Mining, Information Retrieval & $270$ ($2.9\%$)\\\hline
Software Engineering, System Software, Programming Language & $205$ ($2.2\%$)\\\hline
Computer Architecture, Parallel and Distributed Computing, Storage System & $196$ ($2.1\%$)\\\hline
Network and Information Security & $162$ ($1.7\%$) \\\hline
Computer Science Theory & $123$ ($1.3\%$)\\\hline
Human-Computer Interaction and Pervasive Computing & $86$ ($0.9\%$) \\\hline
Uncertain Topics & $5,984$ ($64.2\%$) \\\hline
Total & $9,321$ \\
\bottomrule
\end{tabular}
\label{all domain distribution}
\vspace{-4mm}
\end{table}

\subsection{Citation Graph}
\begin{figure}[!t]
    \centering
    \includegraphics[width=0.5\textwidth]{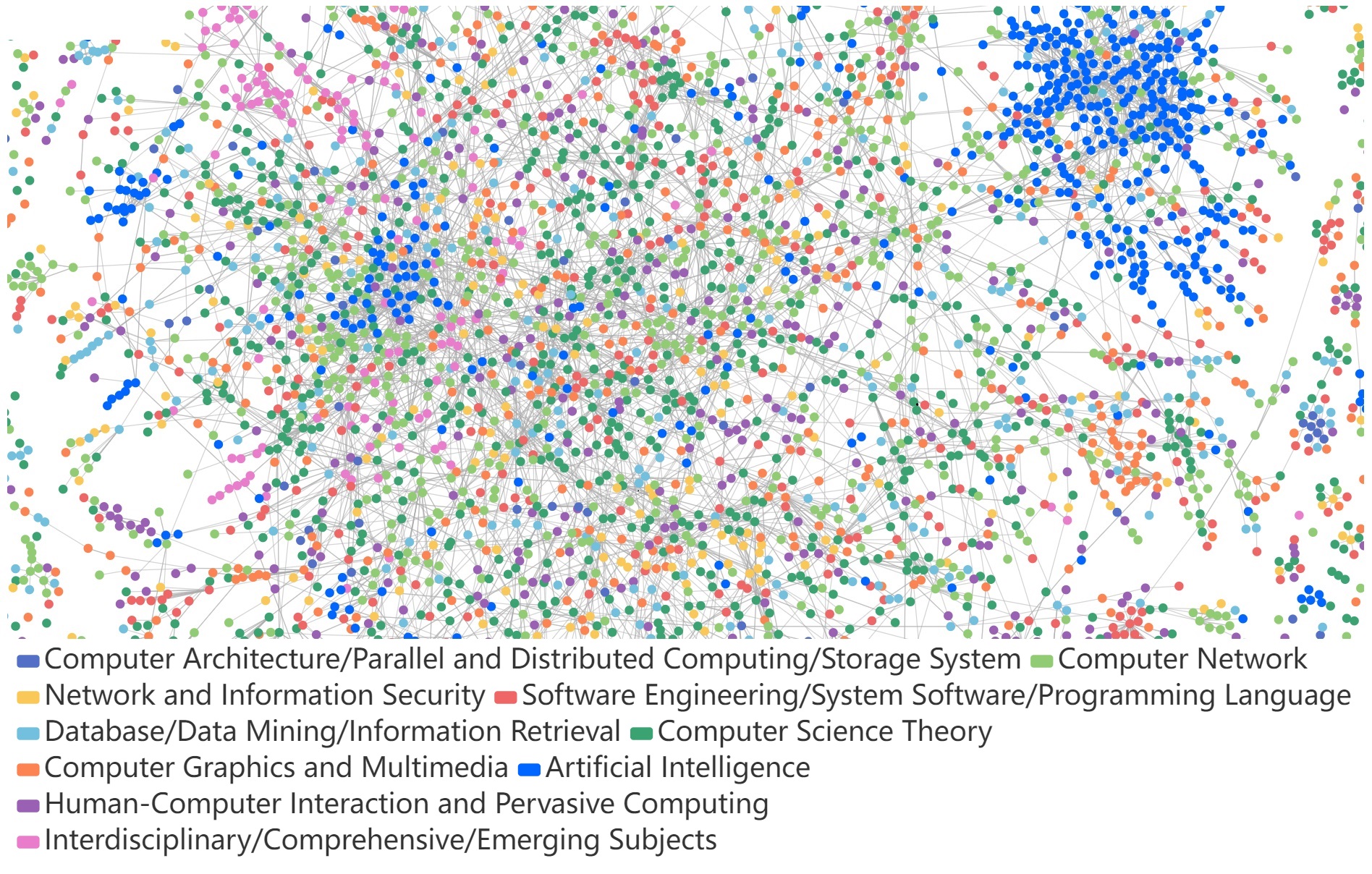}
    \caption{A citation graph which consists 10,000 papers sampled from our whole citation graph. Different colors indicate different subjects in the domain of computer science.}
    \label{fig:citation-graph}
    \vspace{-5mm}
\end{figure}

We initially collect around 80 million scientific papers covering all kinds of subjects such as biology, chemistry and computer science from S2ORC. Then we pick all papers in the domain of computer science by the provided attribute ``domain" to get around 6 million computer science papers. These papers would be linked by citation relationship and publish time to form citation graph. We also visualize a part of connected citation graph consisting of 10,000 papers randomly sampled from our whole citation graph as shown in Fig.~\ref{fig:citation-graph}.   

\section{Our Approach of RePaGer system} \label{sec:model}
In this section, we first introduce the overall architecture of our RePaGer system and then present a detailed description of the Node-Edge Weighted Stiner Tree
(NEWST) model for path selection over a citation graph.

\begin{figure*}[!t]
    \centering
    \includegraphics[width=0.9\linewidth]{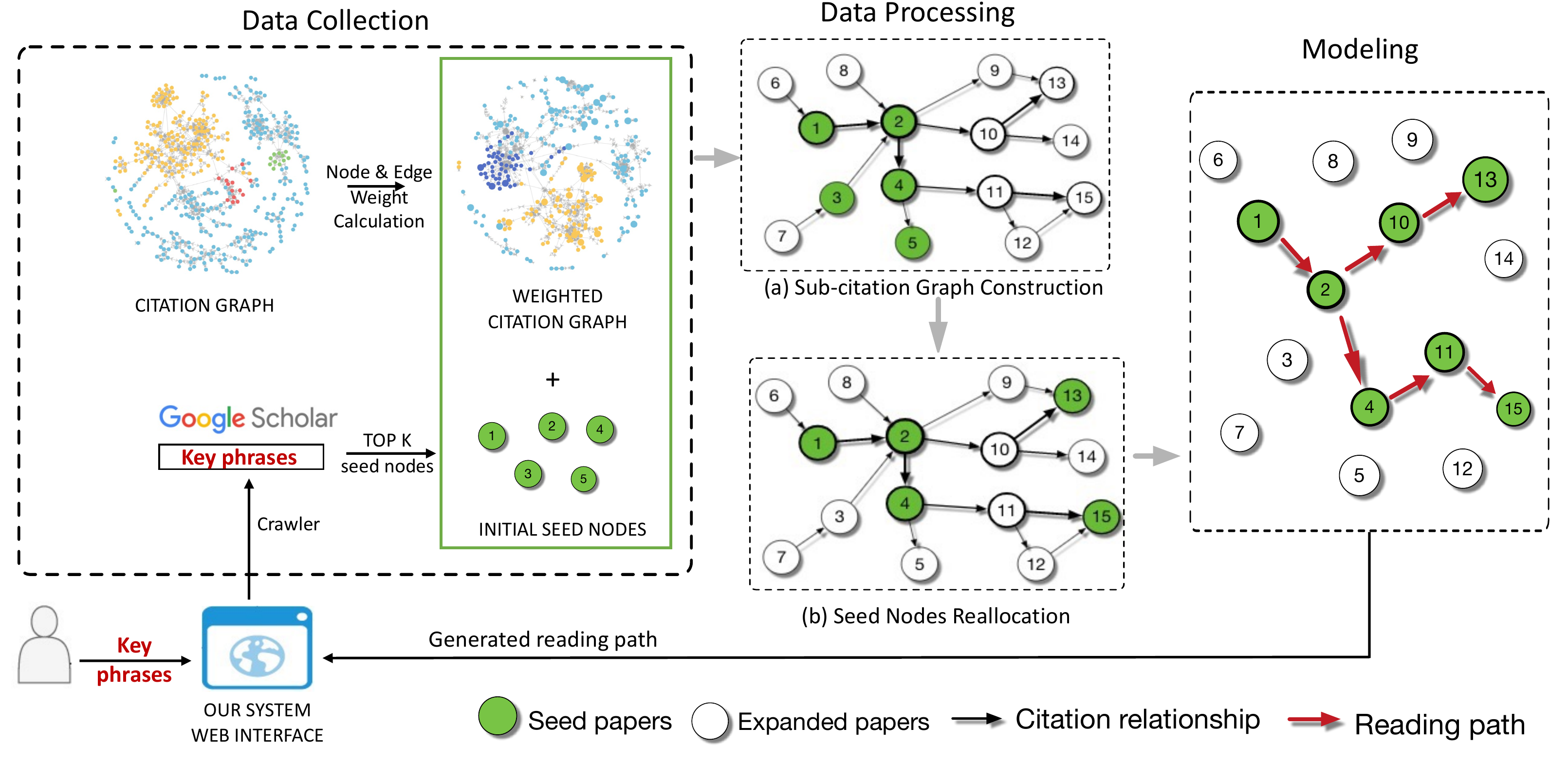}
    \vspace{-4mm}
    \caption{Overall framework of our RePaGer system}
    \label{fig:backend}
    \vspace{-4mm}
\end{figure*}

\subsection{The Architecture of RePaGer System}
Our RePaGer system for Reading Path Generation takes key phrases as inputs and produces a reading path for users to read. Fig. ~\ref{fig:backend} shows the overall architecture of RePaGer. There are five main steps, which would be detailed in the following: 

\begin{enumerate}
\item \textbf{Initial seed nodes obtainment.} 
Our RePaGer system first leverages SerAPI~\cite{serapi} tool to get top $K$ articles from Google Scholar by searching a user's query.

\item \textbf{Weighted citation graph construction.} 
We construct a large-scale citation graph with 6 million papers from S2ORC~\cite{lo-etal-2020-s2orc} in the field of computer science by citation relationship. We further apply our proposed methods on original citation graph to calculate the weight of each paper. The weights of nodes and edges will be introduced in Section \ref{subsec:newst-algorithm}.

\item \textbf{Sub-citation graph construction.} 
After acquiring the initial seed papers from Google Scholar and constructing the weighted citation graph, we capture the first-order and second-order neighbors of initial seeds as candidates to form sub-citation graph with weighted edges and nodes based on the observation that most of the papers relevant to the query can be found among the first-order and second-order of initial seeds as shown in Fig.~\ref{fig:stats-comparison}.

\item \textbf{Seed nodes reallocation.} 
Query's prerequisite chains hidden in citation network cannot be found in initial retrieved results from Google Scholar.
In order to capture more preconditions of the topic and generate a logical reading path with prerequisite chains, we reallocate the initial seed papers by selecting articles with high co-occurrence as new seeds in weighted citation graph. For example, as in Fig.~\ref{fig:backend}, paper 13 co-occurs in the reference list of paper 9 and paper 10, and then paper 13 would be considered as a new seed node because of high co-occurrence. We infer that these new selected papers are more likely to be prerequisite articles for understanding a topic because multiple articles directly relevant to this topic mention them in their papers.

\item \textbf{Reading path generation by NEWST.} 
The NEWST is a graph optimization-based approach to address the Reading Path Generation problem.
Given a weighted citation graph and a subset of vertices in this graph as seed papers (the result of step 4), the NEWST spans through the weighted citation graph, and finds an optimal tree interconnecting all papers in this subset at the minimum cost. Our algorithm not only considers how important each individual paper is in the whole graph, but also takes into account the edge weights between papers. 
We will describe more details about the NEWST in the following section.
\end{enumerate}

\subsection{The NEWST MODEL}
\label{subsec:newst-algorithm}

Scientific paper citation networks usually consist of massive papers, and the prerequisite concepts of a given query are hidden in them. 
In this regard, we propose Node-Edge Weighted Steiner Tree (NEWST) model to identify a reading path for a given query.
Note that our proposed system and algorithm is not only suitable for Reading Path Generation in the filed of computer science or other domains, but also easy to transfer to solve other weighted graph related problems to discover useful information hidden in the graph.
We also give a heuristic solution for NEWST.
The classical node-weighted Steiner tree problem \cite{segev1987node,duin1987some,sun2017node} is defined as: given a graph $G=(V, E)$ with node weight $w: V \rightarrow R^+$ and a subset $S \subseteq V $, find a spanning tree which contains all the nodes in $S$ with minimum overall cost. 


Let $G=(V, E, S, w, c)$ be a connected and undirected network / graph, where $V$ is the set of vertices, $E$ is the set of edges, $w$ is a function which maps each vertex in $V$ to a positive vertex weight, and $c$ is a function which maps each edge from $E$ to a positive edge cost. Let $S\subseteq V$ be a subset of $V$ called compulsory terminals in classical node-weighted Steiner tree problem. The compulsory terminals in our model are reallocated seed papers we retrieve from Step 3 Seed Papers Reallocation.
The purpose of this task is to find an optimal Steiner tree $T$ that spans $S$ with minimal total cost on its edges and nodes, and $T$ is defined below:
\begin{equation*}
T=(V_T, E_T),  S\subseteq V_T\subseteq V, E_T\subseteq E
\end{equation*} 
which minimizes the objective function $\textit{cost}$ of our whole model
\begin{align} \label{steiner_tree_objective}
minimize \quad \textit{cost}(T) = \sum_{ e \in E_T} c(e) + \sum_{ v \in V_T} w(v),
\end{align}
where $e$ is an edge and $v$ is a vertex in the optimal Steiner tree $T$.

The edge cost function $c$ in Eq. \eqref{steiner_tree_objective} is defined as:
\begin{align}
c(i, j) = \frac{\alpha}{con(i, j)^{ \beta} }, 
\end{align}
where $i$ and $j$ are indexes of two different papers, $\alpha$ and $\beta$ are positive constant values, and $con(i, j)$ is a score that measures the relevance between paper $i$ and paper $j$. In our model, $con$ is calculated by how many times paper $j$ cited in paper $i$ or inversely. 

The vertex cost function $w$ in Eq. \eqref{steiner_tree_objective} is defined as:
\begin{align}
w(i) = \frac{ \gamma}{a*pgscore(i) +  b*venue(i)},
\end{align}
where $i$ is a paper$'$s index, $\gamma$ is a positive constant value, $pgscore(i)$ denotes the PageRank \cite{page1999pagerank} score of paper $i$ in our scientific citation network, and $venue(i)$ is the paper $i's$ venue score that we calculate according to venue rankings from AMiner and CCF. This comprehensive venue rankings span 10 topics with a total of around 700 top journals and conferences in the field of computer science. This collection of journals and conferences are manually divided into three levels of rankings by experts in this area in CFF while AMiner also gives an influence score automatically for each of them based on the citations of the best papers in each journal or conference. We calculate the average of the two as a final venue score of each paper in our model;
$a$ and $b$ are positive constant values.
Note that we can revise the definition of the cost function of edges and weight function of nodes to incorporate more valuable information for generating a better reading path.
For example, instead of only exploit the title and citation relationship in our model, we can further utilize the semantic information of the main text. We leave these to our future work.

\begin{algorithm}[t] 
\caption{Heuristic Solution for NEWST.}
\label{alg:newst}
\begin{algorithmic}[1]
\Require
  A connected undirected node-edge weighted graph $G=(V, E, S, w, c)$, with compulsory terminals $S\subseteq V$, vertex cost function $w$ and edge cost function $c$;
\Ensure
  An optimal Steiner tree $T=(V_T, E_T)$ containing all nodes in $S$,   $S\subseteq V_T\subseteq V, E_T\subseteq E$;
\State Construct the complete undirected distance graph  $G_1=(V_1, E_1, S, w, c)$ from $G$;
\State Find the minimum spanning tree, $T_1$, of $G_1$. (If there are several minimal spanning trees, pick an arbitrary one.);
\State Construct the subgraph, $G_s$, of $G$ by replacing each edge in $T1$ by its corresponding shortest path in $G$. (If there are several minimal spanning trees, pick an arbitrary one.);
\State Find the minimal spanning tree, $T$, of $G_s$. (If there are several shortest paths, pick an arbitrary one.) Then $T$ is our optimal Steiner tree containing all nodes in $S$;
\end{algorithmic}
\label{algo:heuristic}
\end{algorithm}

\textbf{Heuristic Solution for NEWST}. Since the classical Steiner tree problem in graphs is NP-hard, and our \textit{NEWST} is a more general version of the Steiner tree, therefore the NEWST problem is also NP-hard, which means that there may not be an algorithm that can solve the problem with a large number of instances in polynomial time. 
One common solution to deal with large network inefficiency is a heuristic algorithm, which finds sub-optimal solutions in large networks in a short time.
In NEWST, we leverage a heuristic algorithm \cite{kou1981fast} to solve this problem and briefly summarize it in Algorithm \ref{algo:heuristic}. The general idea is that the minimum of NEWST can be approximated by computing the minimum spanning tree of the subgraph of the metric closure of $G$ induced by compulsory nodes $S$, where the metric closure of G is the complete distance graph in which each edge is weighted by the shortest path distance between the nodes in $G$.
A shortest path from paper $P_i$ to $P_j$ is a path from $P_i$ to $P_j$ whose distance, including node costs and edge weights, is minimal among all the possible paths from $P_i$ to $P_j$. The minimum spanning tree of a graph is a subset in this graph where the total distance on its edges and nodes is minimal among all spanning trees. As described in Algorithm \ref{algo:heuristic}, the worst case time complexity of Step 1-4 is $O(|S||V|^2)$, $O(|S|^2)$, $O(|V|)$ and $O(|V|^2)$, respectively. Hence, our heuristic algorithm has a worst case time complexity of $O(|S||V|^2)$ and guarantees to output a tree that spans $S$ with total cost on its nodes and edges no more than $2(1-\frac{1}{l} )$ times of the global optimal value, where $l$ is the number of leaves in the optimal Steiner tree $T$. 

\section{User Interface of RePaGer System}

\begin{figure*}[!t]
    \centering
    \includegraphics[width=160mm]{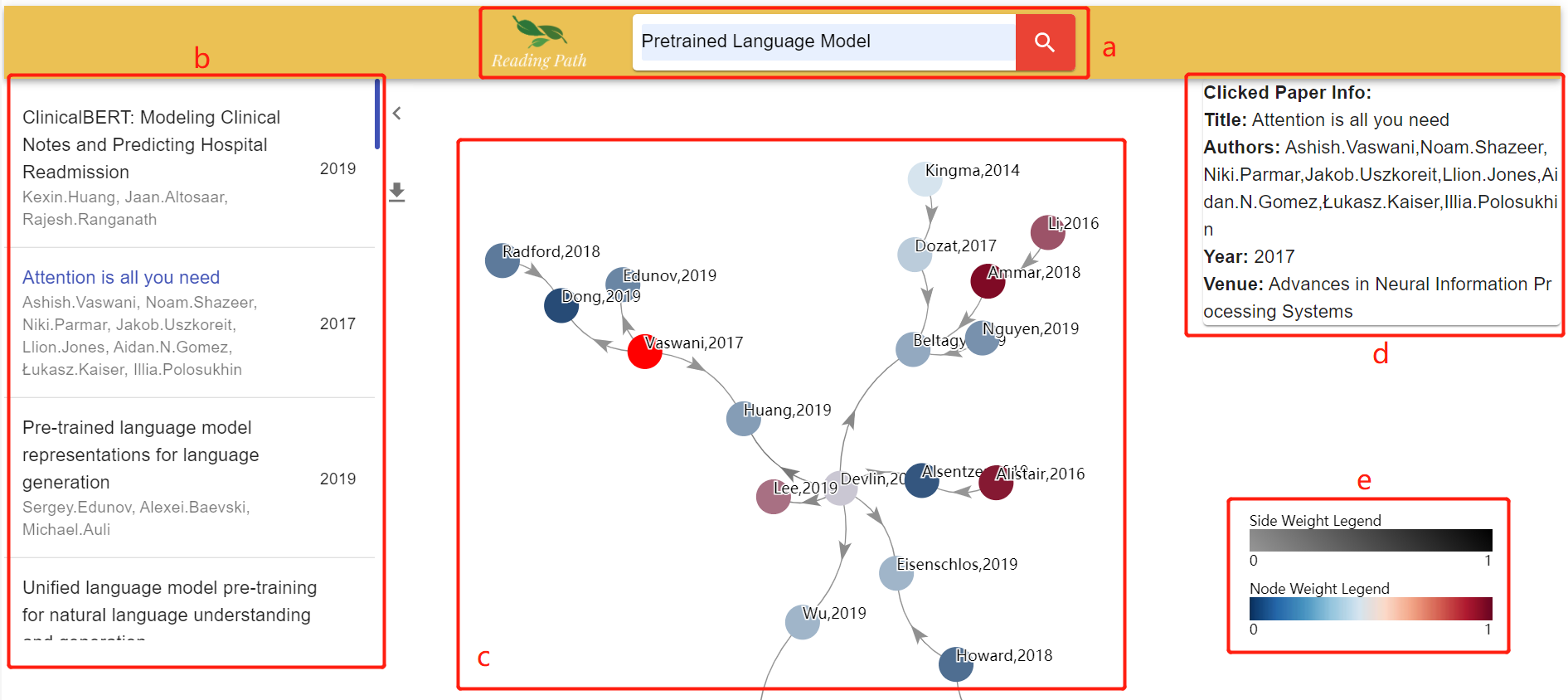}
    \vspace{0mm}
    \caption{User interface of RePaGer system. Panel (a), (b), (c), (d) and (e) indicate ``input panel", ``navigation bar", ``generated reading path  panel", ``detailed information of the clicked paper" and ``node and edge weight annotation" respectively. The content shown in the figure is a reading path generated for the query of ``Pretrained Language Model". }
    \label{fig:interface}
    \vspace{0mm}
\end{figure*}

The user interface of \textbf{Re}ading \textbf{Pa}th \textbf{Ge}ne\textbf{r}ation (\textbf{RePaGer}) system is a web application with input panel same as normal academic search engines like Google Scholar.
An user can input an arbitrary string for an interested research topic, which can be a single key phrase or even a combination of multiple key phrases. As shown in Fig. ~\ref{fig:interface}, after an user inputs a query, the web will present five components consisting of input panel (a), navigation bar (b), generated reading path panel (c), detailed information of the clicked paper (d) and node and edge weight annotation (d), which would be detailed in the following.

\begin{itemize}

\item \textbf{Input panel.} 
This allows users for the subsequent query for a new research topic without coming back main UI interface.
\item \textbf{Navigation bar.} 
Users can leverage the navigation bar to browse papers in the generated path in a flattened way. Each paper is provided with meta information including a title, authors and published year.

\item \textbf{Generated reading path panel.} 
In the generated path, the direction between two papers indicates a reading order. The color of each paper and edge in the path denotes the importance of this paper in the whole path and the relevance between two papers respectively.

\item \textbf{Detailed information of the clicked paper.} 
When a paper either in b or c is clicked, more detailed information about this paper would show up in this component.

\item \textbf{Node and edge weight annotation.} 
Side weight legend annotates the importance between two papers, which means how important a cited paper to its citing paper. From left to right of the legend indicates the importance from low to high.
For the node weight legend, it annotates the importance of the paper in a whole reading path.
\end{itemize}

The RePaGer system also provides two ways for users to interact with the generated reading path. One way is to start with navigation bar to browse papers and then click on interested papers. After that, the color of the clicked paper in component c would become red, and users can easily get relevant information of this paper in a whole reading path. As shown in Fig. ~\ref{fig:interface}, the paper of ``Attention is all you need" in component b is clicked and then the corresponding paper in component c gets red.
Another way is to directly click one interested paper in component c, and then corresponding detailed information of the clicked paper should be presented in component d.
\section{EXPERIMENTS}

\subsection{Experiment Setup}
\textbf{Experimental Details} We evaluate the NEWST model on the proposed SurveyBank Dataset. The key phrases extracted from the title of each survey paper are taken as the model input and the true label is reference papers of the corresponding survey. It is worth noting that we remove the corresponding survey paper of input key phrases to avoid data leakage if it appears in the response from Google Scholar. Since the limitation of the publication time for each survey, the ground truth papers cannot cover the newest literature. Therefore, all papers considered during the metric computation are published before the time when the survey was issued.
Following the setting of \cite{sun2017node}, the parameters $\{\alpha, \beta, \gamma, a, b\}$ used in the NEWST algorithm are set as $\{3, 2, 5, 0.7, 0.3\}$. Moreover, we set the number of initial seed papers from Google Scholar to be $30$ for all experiments.

\textbf{Baselines} We compare our method with the following models:
\begin{itemize}
    \item \textbf{Google Scholar:} A paper retrieval engine developed by Google which can return a list of papers according to the input phrases. For each query phrase, we select the top-$K$ papers in the retrieval results as the generated results.
    \item \textbf{Microsoft Academic:} A paper retrieval engine developed by Microsoft. Correspondingly, the top-$K$ retrieved papers of the query phrase are chosen as the final results.
    \item \textbf{AMiner:} Another academic search engine released by Tsinghua University. Similarly, we take the top-$K$ retrieval results in our experiments.
    \item \textbf{PageRank \cite{page1999pagerank}:} A well-known Web page ranking algorithm which orders pages based on their citation relationship. Similar to the NEWST, we first expand initial seed nodes returned from Google Scholar to their neighbors as candidates, and then the PageRank algorithm is applied to reorder initial seeds and expanded candidates together. 
    \item \textbf{SciBERT \cite{beltagy2019scibert}:} An awesome variant of the BERT model \cite{devlin2018bert} trained on scientific literature from the corpus of Semantic Scholar \cite{semantic}. We train a matching model using SciBERT to score the matching degree of queries with paper titles and abstracts. During the inference phase, we also expand the seed nodes returned from Google Scholar and then re-rank them via our trained matching model.
\end{itemize}

\begin{figure*}[!t]
\centering
\resizebox{.85\textwidth}{!}{
\begin{subfigure}[t]{0.3\textwidth}
\centering
  \includegraphics[width=\textwidth]{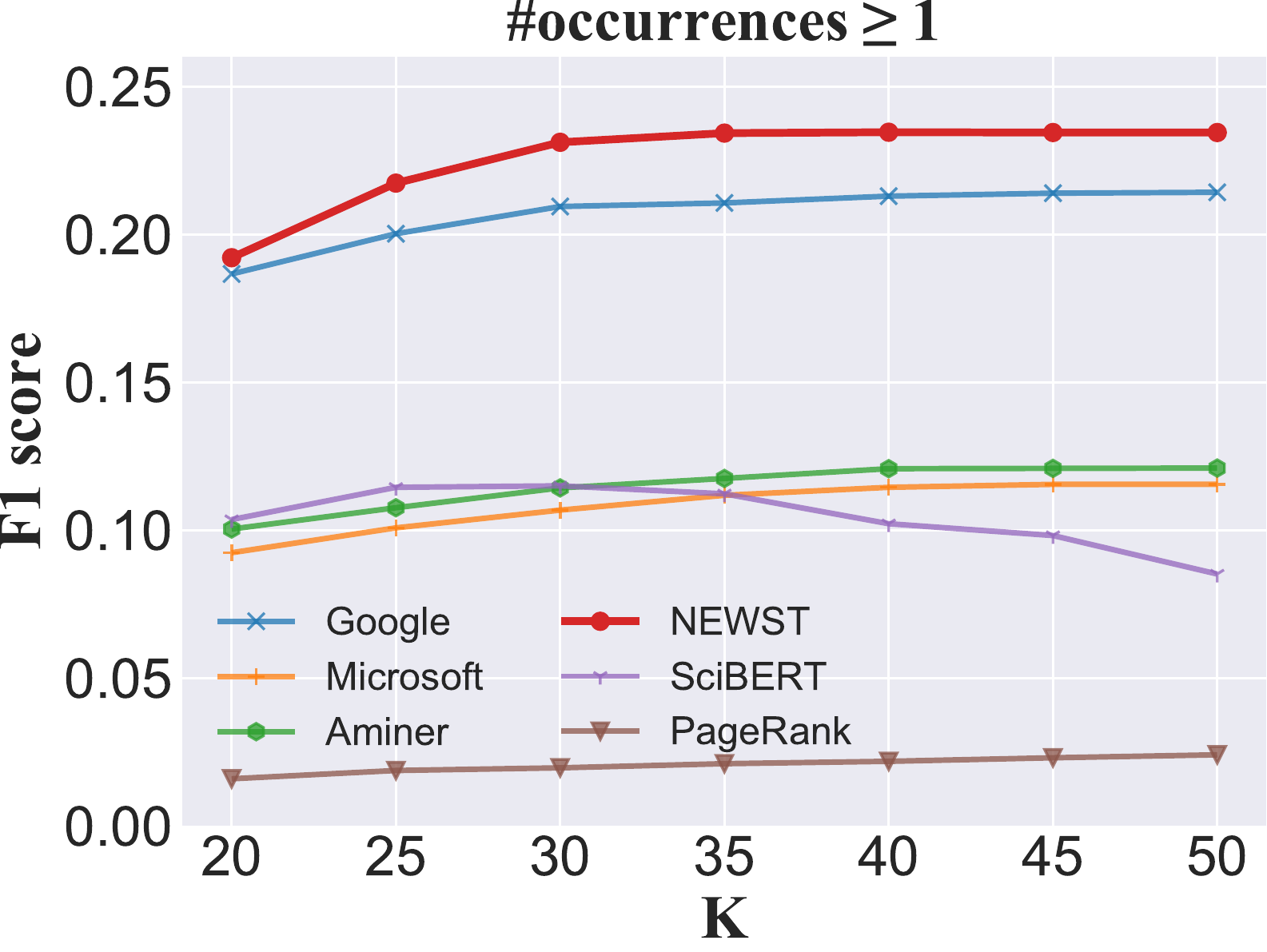}
\end{subfigure}
\begin{subfigure}[t]{0.3\textwidth}
\centering
  \includegraphics[width=\textwidth]{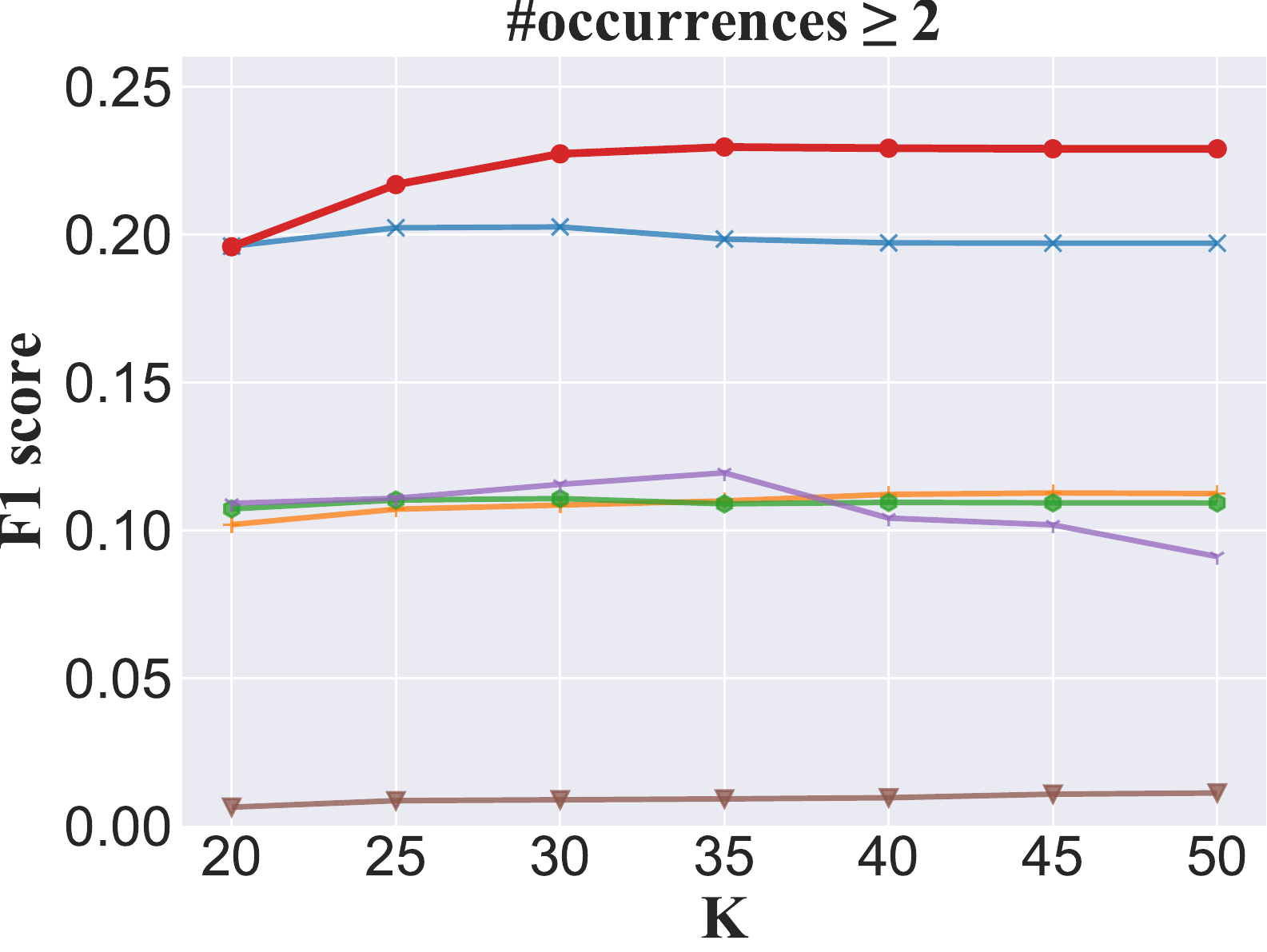}
\end{subfigure}
\begin{subfigure}[t]{0.3\textwidth}
\centering
  \includegraphics[width=\textwidth]{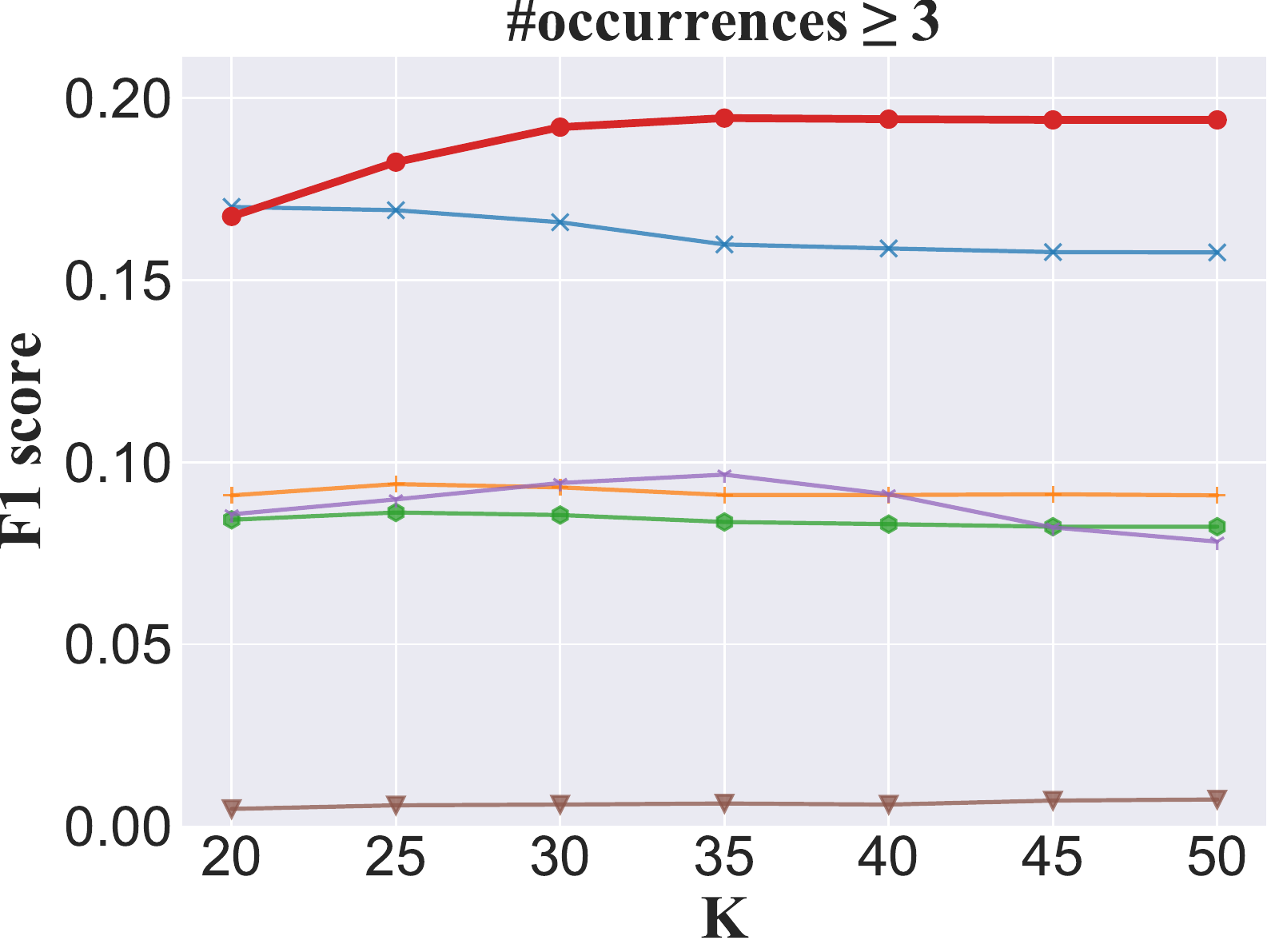}
\end{subfigure}}
\resizebox{.85\textwidth}{!}{
\begin{subfigure}[t]{0.3\textwidth}
\centering
  \includegraphics[width=\textwidth]{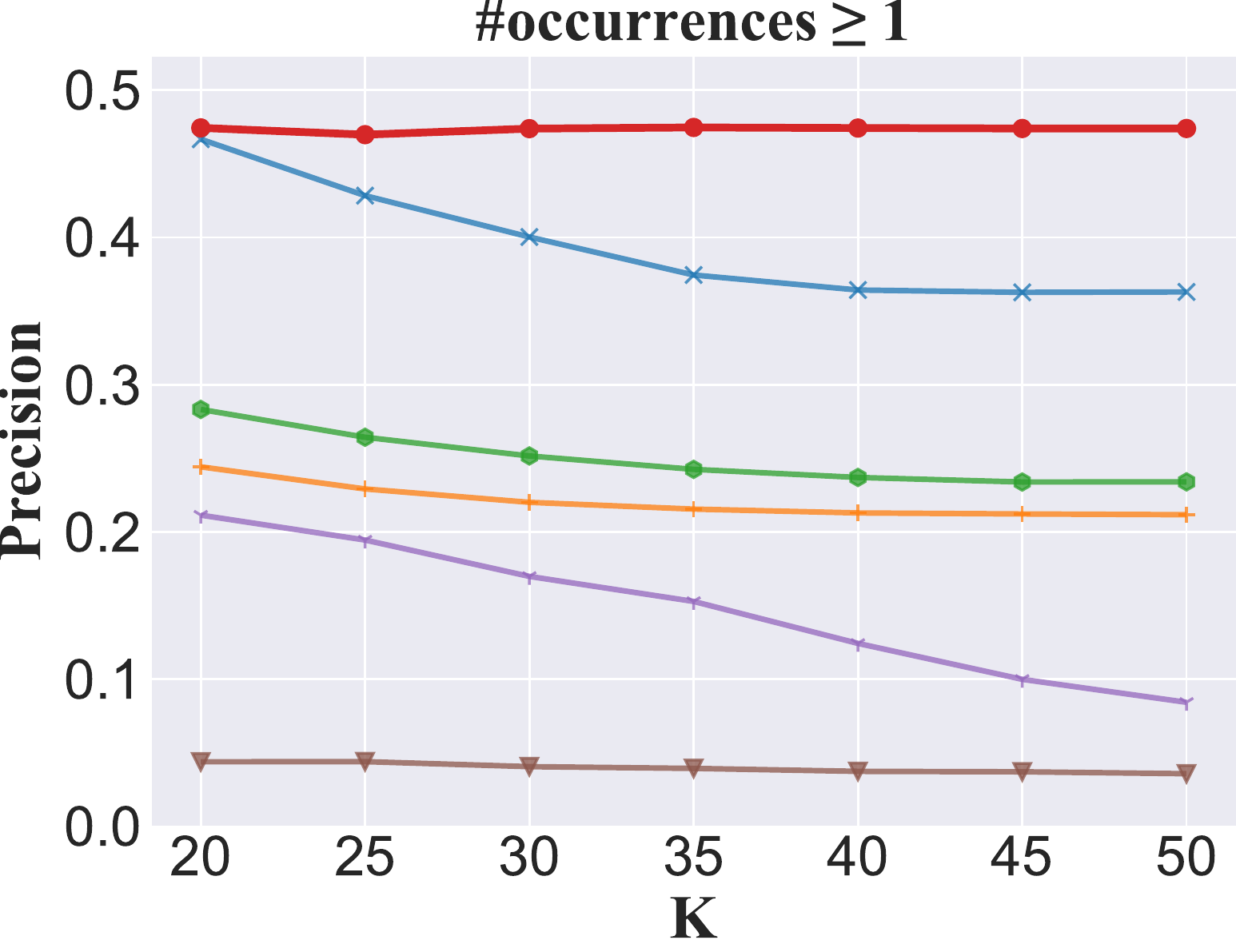}
\end{subfigure}
\begin{subfigure}[t]{0.3\textwidth}
\centering
  \includegraphics[width=\textwidth]{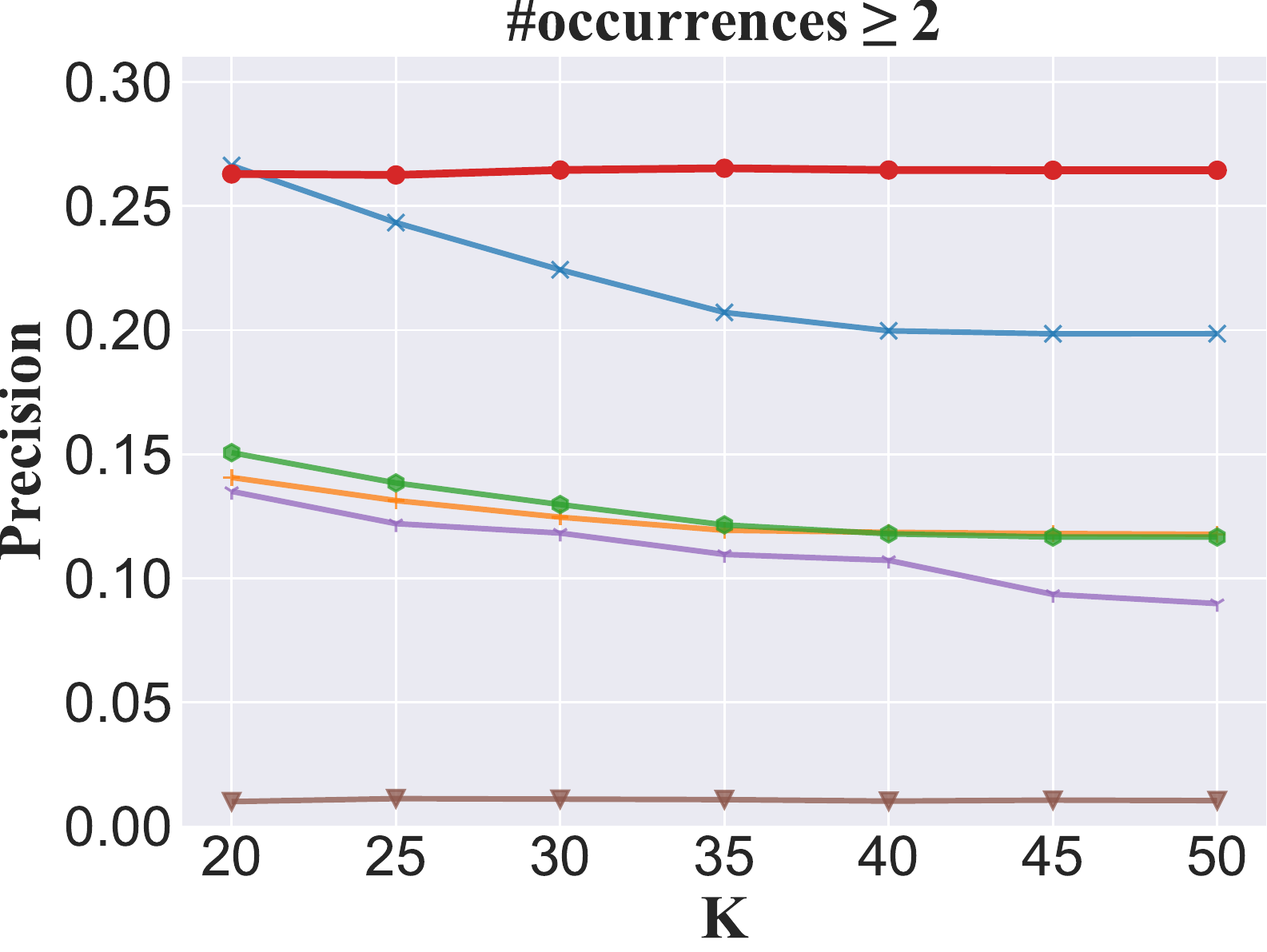}
\end{subfigure}
\begin{subfigure}[t]{0.3\textwidth}
\centering
  \includegraphics[width=\textwidth]{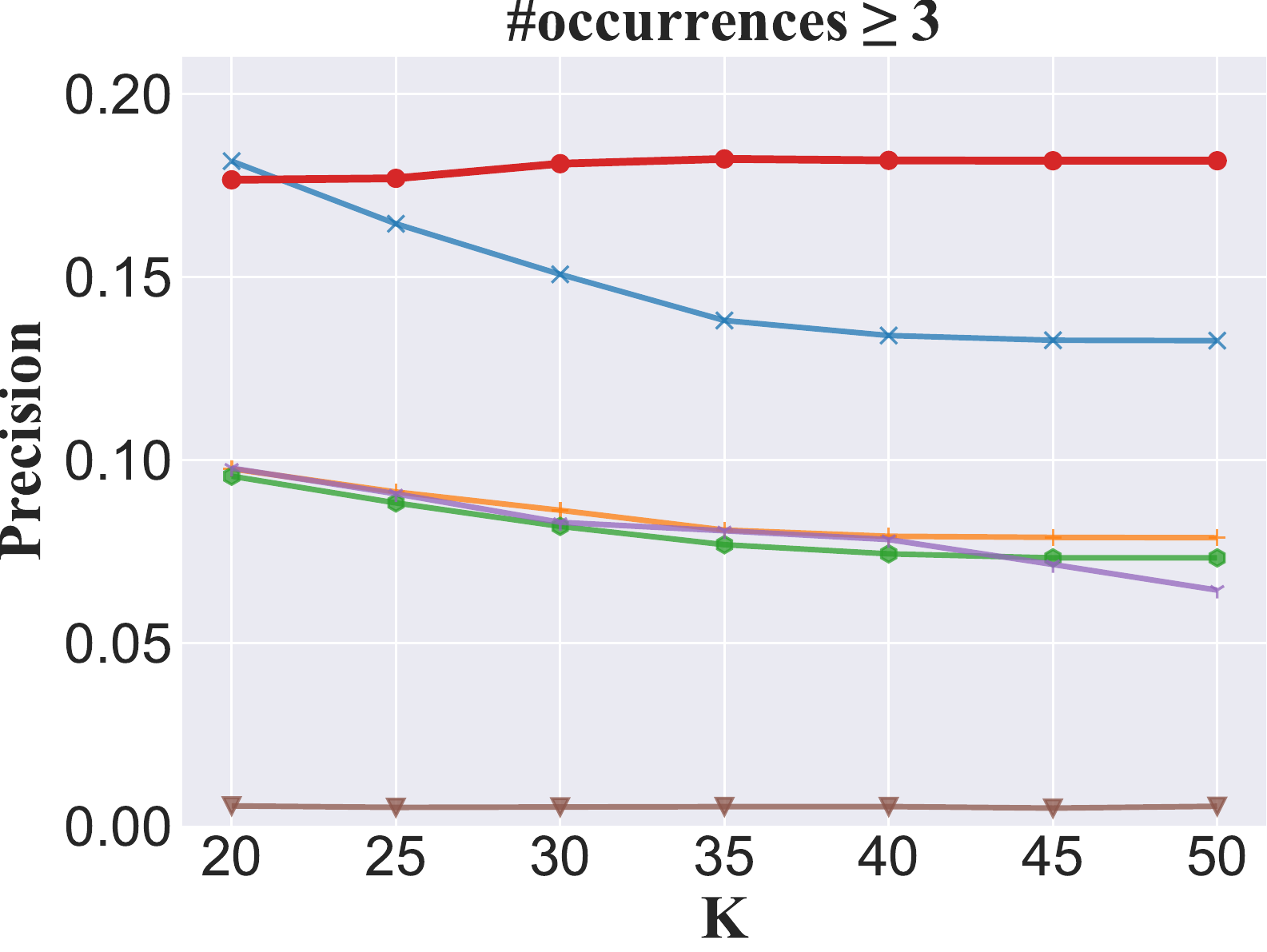}
\end{subfigure}}
\vspace{-2mm}
\caption{The performance comparison of NEWST, Google Scholar, Microsoft Academic, AMiner, PageRank and SciBERT on F1 score and precision metrics for top-$K$ recommended papers.}
\label{fig:diff_apear_time}
\vspace{-4mm}
\end{figure*}

\textbf{Task Evaluation}
As described in Section \ref{sec:task-defi}, we combine metric evaluation and human evaluation to estimate the quality of generated reading paths.
On one hand, we first downgrade the evaluation of reading paths to the evaluation of reading lists. The reason is that \textbf{once the reading list is determined, the reading direction between two papers can be easily and uniquely obtained from our constructed citation graph based on citation relationship and published time}. Therefore, the quality of Reading Path Generation solely depends on the generated reading lists.
As our model returns a list of candidate papers for each input query phrase, we adopt P@$K$ (precision) and F1@$K$ (F1 score) to evaluate the performance of our model, where $K$ denotes the number of generated papers. The results averaged over all query phrases are reported as the final performance.
On the other hand, we further perform human evaluation on the generated reading paths to further evaluate their quality. 

\subsection{Overlapping Metric Evaluation}

\textbf{Overall Performance}
To test the performance of our model, we compare it with several outstanding methods on the task of generating a list of top-$K$ papers, where $K$ ranges from $20$ to $50$ as each survey at least cites 20 papers. The results are shown in Fig. \ref{fig:diff_apear_time} in which we aim to generate the ground truth papers that appear in the survey paper with different times.   From these experimental results, it can be seen that for almost all evaluation scenarios, our proposed NEWST model outperforms all other methods, especially when the number of generated papers becomes large. In particular, when the number is large than 25, NEWST consistently outperforms the academic search engine (i.e., Google Scholar, Microsoft Academic and AMiner) by a substantial margin. This is caused by the fact that existing search engines solely return the paper whose title contains query phrases.

It is worth noting that by exploring the citation relationship among literature, NEWST performs significantly better, compared with SciBERT based matching model, which plainly employs semantic analysis, highlighting the importance of using the Steiner tree algorithm to discover high-order reference chain. Although introducing citation information may attribute to the quality of generating reading list, solely considering this type of information is inclined to deteriorate the performance. This phenomenon can be clearly observed in Fig. \ref{fig:diff_apear_time}, in which the PageRank method achieves the worst results, as it always returns the papers whose citation number is the largest, which is fallacious since in the real world scenarios, the most relevant papers may not be the most cited papers. The proposed NEWST model relieves this issue by introducing the node-edge weighted Steiner tree algorithm and, narrows down the candidate papers via seed nodes selection and attains significant performance gains. Another interesting observation is that NEWST preserves superior performance no matter $K$ is large or small, especially in terms of precision. For large $K$, the performance of all baseline methods generally tends to drop significantly, while our model achieves even improved precision, indicating that our method is more robust and more suitable for learners who like to acquaint more relevant knowledge.

\textbf{Comparison of Different Seed Paper Numbers}
To investigate the impacts of the seed nodes selection, we conduct a parameter sensitive study with different numbers of seed nodes and the performance results are demonstrated in Table \ref{table:diff_num_seed_nodes}. It can be observed that the performance of our model is still robust even though the node number varies in a large interval. Specifically, increasing the seed node number, the performance rises steadily. The reason for this expected phenomenon lies in the fact that the more seed nodes to be used, the more ground truth papers can be included after expansion, highlighting the importance of considering multi-order citation relationship among the papers returned from Google Scholar, which solely takes into account of the keywords in user queries. However, overloading the quantity decreases the performance of precision, which might be caused by introducing ineffectual noise paper when using too many seed nodes.



\textbf{Effect of Seed Papers Reallocation} We experiment with four variants for our model to investigate the effect of seed papers reallocation: \romannum{1}) NEWST: with high co-occurrence papers as compulsory nodes for the NEWST; \romannum{2}) NEWST-W: with initial top-30 seed papers as compulsory nodes; \romannum{3}) NEWST-U: with the union of above two; \romannum{4}) NEWST-I: with the intersection of above two.   Table \ref{table:ablation_study} presents the results of all variant methods. Comparing NEWST with NEWST-W, it can be observed that the injection of seed papers reallocation has improved  the robustness of generating reading list, resulting in better performance. Moreover, NEWST-I, which uses the intersection of initial seed papers and co-occurrence papers as compulsory nodes, achieves comparable results with NEWST, further confirming the superiority of considering co-occurrence times during reallocating the compulsory nodes for the NEWST algorithm. However, by using the union of them, NEWST-U achieves even better results in F1 score but worse in precision. This may be caused by that the union of papers increases the number of compulsory nodes, which results in including more ground truth papers such that rising up the recall rate, but also introducing noise to reduce the precision.

\textbf{Impact of the Node-Edge Weight} To futher evaluate the impacts of node and edge weights, we propose the following variant methods: \romannum{1}) NEWST-C: method that uses reallocated papers as final results (i.e., ignoring step 4); \romannum{2}) NEWST-N: excluding node weights in Algorithm \ref{alg:newst} and \romannum{3}) NEWST-E: excluding edge weights. As shown in Table \ref{table:ablation_study}, by introducing node and edge weights, NEWST-N, NEWST-E, and NEWST consistently outperform NEWS-C in terms of F1 score, which means that the proposed Steiner tree algorithm can effectively involve the ground truth papers in the generation of reading path. Compared with NEWST-C, although the exploitation of NEWST algorithm may damage precision, our model still achieves comparable performance. More importantly, due to the inability of path generation, NEWST-C cannot take account of the prerequisite relationship. However, the proposed NEWST model not only tells users ``what to read'', but also ``how to read''.

\subsection{Human Evaluation}
To further examine the quality of our generated reading path, we conduct a human evaluation of Google Scholar and our RePaGer System.
We conducted a human evaluation to examine the quality of the generative reading path. The following is a detailed description of the evaluation procedure.

\textbf{Datasets} We used the proposed SurveyBank dataset in human evaluation. Specifically, we randomly selected 20 query samples from the domain of Artificial Intelligence and Information Retrieval, respectively. As shown in Fig. \ref{fig:interface}, for each query sample, the reading path and reading list retrieval results were generated by RePaGer system and Google Scholar, respectively. These retrieval results and query sentences were used as evaluation materials to compare the usability of the two systems.


\begin{table}[!t]
	\caption{Impact of the number of seed nodes on the performance of NEWST model.}
	\centering
	\setlength{\tabcolsep}{0.85mm}{
		\begin{tabular}{c|cccccccc}
		\toprule
		\#seed nodes & 10 & 15 & 20 & 25 & 30 & 40 & 50\\
		\midrule
		F1 score & 0.1924 & 0.2094 & 0.2197 & 0.2251 & 0.2343 & 0.2385 & $\mathbf{0.2387}$ \\
		Precision & 0.4279 & 0.4454 & 0.4565 & 0.4668 & 0.4743 &  $\mathbf{0.4755}$ & 0.4658 \\
		\bottomrule
	\end{tabular}}
	\vspace{-3mm}
	\label{table:diff_num_seed_nodes}
\end{table}

\textbf{Participants} We recruited 16 participants to evaluate our system. All of them are graduate students of computer science, mainly majoring in  artificial intelligence (AI) or information retrieval (IR). According to their expertise in these two fields, we divided them equally into two groups, and then assigned them to the corresponding evaluation tasks (i.e., 8 persons for 20 queries in AI, 8 persons for 20 queries in IR).

\textbf{Procedure} At the beginning, we briefly introduced the SurveyBank dataset and RePaGer system. Then we take the query phrase ``pretrained language model" as an example, to demonstrate how to conduct evaluation. The retrieval results of RePaGer are shown in Fig. \ref{fig:interface}, respectively. Each participant were asked to explore the recommended papers and think whether these papers are conducive to help them to quickly grasp the subject matters of this field. After careful consideration, the participants filled out a self-evaluation questionnaire and began evaluating the selected 20 query samples.

\textbf{Self-evaluation questionnaire} In the questionnaire, we asked the participants to self-rate their preference of the systems from three folds: \romannum{1}) prerequisite: whether the retrieval results contain prerequisite relationship, i.e., the system should not only tells the reader ``what to read", but also ``how to read"; \romannum{2}) relevance: whether the retrieval results are consistent with the query phrase; \romannum{3}) completeness: whether the retrieval results contain comprehensive knowledge of the query domain. Further, the participants are asked to rank them in three folds: \romannum{1}) prefer the PaReGer; \romannum{2}) prefer the Google Scholar; or \romannum{3}) prefer the two systems equally. The ratings averaged over all participants are reported as final performance.

\textbf{Evaluation results} The results of human evaluation are shown in Table \ref{table:human_evaluation}. Thanks to the incorporation of citation relationship, the generated reading path of NEWST contains more {\em prerequisite} chains, indicating that our model can generate a more reasonable structured reading list. The superiority of our model can be further observed by that the {\em consistency} of the two models is basically the same. However, the comparison on the {\em completeness} shows that NEWST can provide readers more comprehensive knowledge, which further confirms that our model can help users to comprehend this topic profoundly.

\begin{table}[!t]
	\caption{The performance of variant NEWST models validated in ablation study.}
	\centering
	\setlength{\tabcolsep}{0.85mm}{
		\begin{tabular}{c|cc||c|cc}
		\toprule
		Methods & F1 score & Precision & Methods & F1 score & Precision  \\
		\midrule
		NEWST & 0.2343 & $\textbf{0.4743}$ &  NEWST & $\textbf{0.2343}$ & 0.4743   \\
		NEWST-W & 0.2269 & 0.4533 & NEWST-C  & 0.2064 & $\textbf{0.5080}$  \\
		NEWST-I & 0.2345 & 0. 4738 & NEWST-N  & 0.2195 &  0.4978 \\
		NEWST-U & $\textbf{0.2462}$ & 0.4188 & NEWST-E & 0.2203 & 0.4710   \\
		\bottomrule
	\end{tabular}}
	\label{table:ablation_study}
	\vspace{-2mm}
\end{table}

\subsection{Empirical Study of Computational Efficiency}
\begin{table}[!t]
	\caption{Running time of our model under different retrieval cases.}
	\centering
	\setlength{\tabcolsep}{2mm}{
		\begin{tabular}{c|c|c||c}
		\toprule
		& \#nodes & \#edges & Time (seconds)\\
		\midrule
		Case 1 & 851 & 1313 & 9.28 \\
		Case 2 & 1276 & 2208 & 27.99 \\
		Case 3 & 1495 & 2235 & 35.31 \\
		Avg. (test set) & 1754 & 2938 & 60.42 \\
		\bottomrule
	\end{tabular}}
	\label{table:time_efficiency}
	\vspace{-3mm}
\end{table} 
To show the computation efficiency of our proposed method, we also report the running time among several retrieval cases on a machine with 2.40GHz Intel(R) Xeon(R) 14 CPU cores and 132 GB memory. The results are demonstrated in Table \ref{table:time_efficiency}, where \#nodes and \#edges represent the number of nodes and edges in the citation graph constructed in Step 2 of our method, respectively. It can be seen that because of the extra work to generate the reading path from the citation graph, the proposed model is a little bit more time-consuming than the general research engines, but delivers the final results around 1 minute on average. For the readers looking for literature on their own, our model effectively tailors the learning order for them, greatly saving their learning time.

\subsection{Reading Path Visualization}
\begin{figure*}[t!]
\centering
\includegraphics[width=\linewidth]{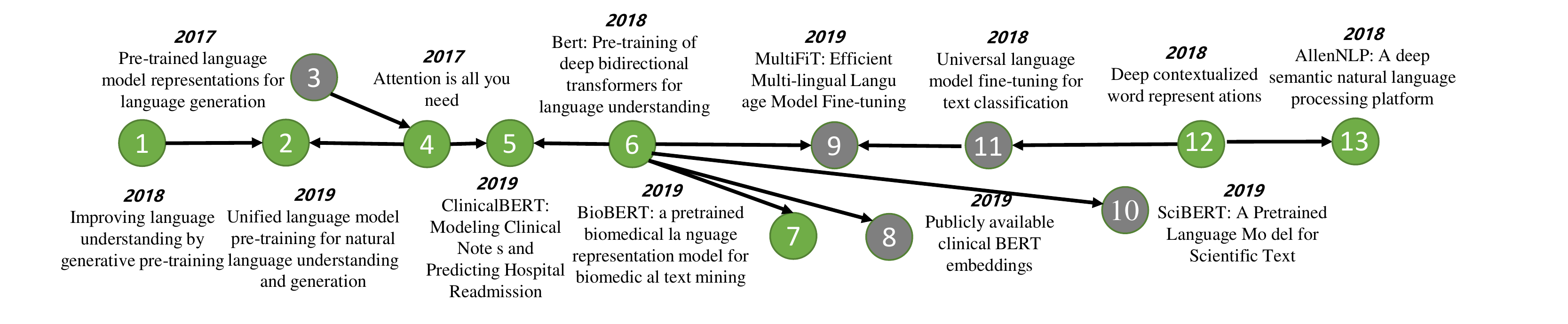}
\vspace{-6mm}
\caption{The reading path generated by our NEWST model in terms of query phrase ``Pretrained Language Model'', where the reading order is determined by arrows. Each green circle denotes a paper not showing up in the TOP 30 papers from Google Scholar while the gray one denotes the paper appearing in the TOP 30 papers from Google Scholar.}
\label{reading_path_visualization}
\vspace{-4mm}
\end{figure*}

Figure \ref{reading_path_visualization} illustrates the reading path of query “Pretrained  Language Model”, where each arrow indicates a reading order between two papers. Due to the space limitation,the reading path demonstrated in Fig.~\ref{reading_path_visualization} is a subtree of the result returned from our RePaGer system. As we can see, by introducing the prerequisite structure into reading path, learners can easily grasp the technological evaluation of a new field and learn the main concept quickly. More importantly, our system not only teaches users “how to  read”, but also tells them ``how to understand". This desirable virtue can be observed in the 4-th, 6-th and 12-th nodes of the reading path. These pre-requisite concepts of Attention, BERT and Contextualized Word Embedding are less directly relevant to the topic of ``Pretrained  Language Model" but essential to help understand technology evolution for the topic profoundly. However, these pre-requisite concepts can not be found in the retrieved TOP list from Google Scholar.

\begin{table}[!t]
	\caption{Human evaluation on the domain of artificial intelligence (AI) and data mining (DM). A and B represent the system of Google Scholar and NEWST, respectively.}
	\centering
	\setlength{\tabcolsep}{1.5mm}{
		\begin{tabular}{c|c|ccc}
		\toprule
		Domain & Criterion & Prefer A (\%) & Same (\%) & Prefer B (\%)\\
		\midrule
		\multirow{3}*{AI}
		& Prerequisite & 0 & 7.90 & 93.10 \\
		& Relevance & 37.93 & 37.93 & 24.14 \\
		& Completeness & 20.69 & 41.38 & 37.93 \\
		\midrule
		\multirow{3}*{DM}
		& Prerequisite & 0 & 23.33 & 76.67 \\
		& Relevance & 26.67 & 43.33 & 30.00 \\
		& Completeness & 20.00 & 36.67 & 43.33 \\
		\bottomrule
	\end{tabular}}
	\label{table:human_evaluation}
	\vspace{-4mm}
\end{table}
\section{Related Works}
With the explosion of the quantity of scientific papers, generating a reading list from a large amount of literature according to the given query phrases has been shown to be essential. Existing research can be roughly divided into two categories: \romannum{1}) reading list generation that ignore the prerequisite relationship among the generated literature, i.e., reading order; \romannum{2}) generating structured reading path which can effectively indicate the readers to learn step by step.

Among the reading order irrelevant methods, the mainstream methods mainly focus on the task of retrieving the most relevant documents for a given query \cite{witten1999managing, wang2007personalized, bogers2008recommending}. Specifically, \cite{tang2008design} introduced the problem of pedagogical paper recommendation which takes account of learner's interest, knowledge, goals, etc, and resolves it by leveraging collaborative filtering techniques \cite{sarwar2001item}. To generate introductory reading lists, \cite{ekstrand2010automatically} explored several collaborative and content-based filtering methods for rating the importance of a paper within the literature. \cite{chang2000learning} proposed another method combined with the gradient decent algorithm to construct customized authority lists. Recently, \cite{jardine2014automatically} proposed to produce reading lists in an unstructured manner in which several relevant documents of the given query phrases are first chosen by the LDA topic model \cite{blei2003latent}, and then a ThemedPageRank approach is further developed to re-rank these topic-related papers. However, these methods neglect the prerequisite chain \cite{li2019should} of the literature, downgrading the problem to item-based recommendation, which abominably hinders the application of the models.

In the pedagogical scenario, learners are likely to spotlight the suitable literature and prioritize them based on the prerequisite relationship. To this end, the reading order relevant methods aim to incorporate prerequisite chain more than solely generating reading lists. Regretfully, due to the frustration of extracting prerequisite relation among documents, there are few efforts invested in this field. Recently, \cite{gordon2017structured} proposed to address this issue by adopting concept graph \cite{gordon2016modeling, fabbri-etal-2018-tutorialbank}, which is used to model the concept dependency among knowledge domains and relative documents. However, the concept dependency mining algorithm used in the construction of concept graph is somewhat crude and inaccurate. Besides, the concept graph is not able to cover different domains. Beyond the document recommendation, some data inherently appears with the prerequisite relationships, like lecture data created in the MOOCs \cite{yu2019course,yu2020mooccube}, as different lectures credited by the same student are time-variant. This notorious problem can be automatically sidestepped by leveraging citation relationship.

The problem coming with the generation of reading lists is how to evaluate its quality. In \cite{gordon2017structured}, 33 experts were recruited to take part in the evaluation. It is no doubt that such a direct method is impractical, time and cost consuming. The reference lists of papers to some extent reflect the professional knowledge in their scientific field. Therefore, some prior works \cite{ekstrand2010automatically, mcnee2002recommending, torres2004enhancing} have been proposed to evaluate the coverage rate of the reference list given a query paper. Inspired by that, we publish a public SurveyBank dataset to employ an offline analysis. 
Different from the aforementioned evaluation methods, in this paper, we opportunely exploit the key phrases extracted from the survey title as evaluation query, and its reference list as the ground truth expected to be generated, which is more reasonable and intuitive.

It is worth to note that the proposed reading path generation problem is somewhat relevant to the exploratory data analysis (EDA) problem \cite{bar2020automatically,parette2008benefits,milo2018next} and the composite item formation and exploration (CIFE) problem \cite{basu2010constructing,roy2011interactive}. EDA aims at building an interactive system for users to allow easy-to-use data exploration and visualization. CIFE studies a problem of effective construction and exploration of large sets of packages associated with a central item. 
At first glance, the targets of EDA and CIFE are to some extent similar to RPG, which focuses on generating a paper reading path for someone who lacks preliminary knowledge of a queried research field. 
However, the methods used in EDA and CIFE cannot be directly applied to RPG. More specifically, recent EDA models generally exploit reinforcement learning to automatically generate the data analysis process \cite{bar2020automatically,parette2008benefits,milo2018next}. These methods cannot be applied in RPG since RPG is not an interactive process. 
Furthermore, to apply CIFE in the context of RPG, we need to build up a huge structure of concepts in the domain, which would be another totally different research topic.

\section{CONCLUSION AND FUTURE WORK}
In this paper, we proposed an effective and efficient academic paper retrieval system, which cannot only tell users ``what to read'', but also guide them about ``how to read'' and helps with ``how to understand''. To this end, we formulated a new task named Reading Path Generation and proposed the first corresponding dataset SurveyBank to facilitate future research. To generate a useful learning path, a reasonable and scalable model was further proposed, in which we first expanded the initial seed papers returned from Google Scholar to construct a weighted citation network, and then reallocated the seed papers by selecting the co-citing papers. Finally, the reading path was generated by using a novel node-edge weighted Steiner tree algorithm.  The effectiveness of our framework was demonstrated with extensive experiments. Since how to define a reasonable reading path for a given topic is an open question, and we propose it based on citation relationship, we are looking forward to more research works on this in the future.

\section{Acknowledgment}
This work was in part supported by the Canada CIFAR AI Chair Program, the NSERC Discovery Grant RGPIN-2021-03115, and the UK Engineering and Physical Sciences Research Council under Grant EP/P011829/1.



\bibliographystyle{IEEEtran}
\bibliography{sample-base}

\end{document}